\documentclass[final]{cvpr}

\usepackage{times}
\usepackage{epsfig}
\usepackage{graphicx}
\usepackage{amsmath}
\usepackage{amssymb}
\usepackage[table,xcdraw]{xcolor}
\usepackage{multirow}
\usepackage{appendix}

\usepackage[pagebackref=true,breaklinks=true,colorlinks,bookmarks=false]{hyperref}

\def\cvprPaperID{2094} % *** Enter the CVPR Paper ID here
\def\confYear{CVPR 2021}

\begin{document}

%%%%%%%%% TITLE
\title{Disentangled High Quality Salient Object Detection}

\author{Lv Tang \quad Bo Li$^{1}$\thanks{Corresponding author and equal contribution to first author. This work was supported by National Natural Science Foundation of China 61906036 and the Fundamental Research Funds for the Central Universities (2242021k30056).} \quad Shouhong Ding$^{1}$ \quad Mofei Song$^{2,3}$ \\
$^{1}$Youtu Lab, Tencent, Shanghai, China \\
$^{2}$The School of Computer Science and Engineering, \\
$^{3}$The Key Lab of Computer Network and Information Integration (Ministry of Education), \\
Southeast University, Nanjing, China \\
{\tt\small luckybird1994@gmail.com}, {\tt\small libraboli@tencent.com}, 
{\tt\small ericshding@tencent.com}, \\{\tt\small songmf@seu.edu.cn}
}

\maketitle

%%%%%%%%% ABSTRACT
\begin{abstract}
Aiming at discovering and locating most distinctive objects from visual scenes, salient object detection (SOD) plays an essential role in various computer vision systems. Coming to the era of high resolution, SOD methods are facing new challenges. The major limitation of previous methods is that they try to identify the salient regions and estimate the accurate objects boundaries simultaneously with a single regression task at low-resolution. This practice ignores the inherent difference between the two difficult problems, resulting in poor detection quality. In this paper, we propose a novel deep learning framework for high-resolution SOD task, which disentangles the task into a low-resolution saliency classification network (LRSCN) and a high-resolution refinement network (HRRN).  As a pixel-wise classification 
task, LRSCN is designed to capture sufficient semantics at low-resolution to identify the definite salient, background and uncertain image regions. HRRN is a regression task, which aims at accurately refining the saliency value of pixels in the uncertain region to preserve a clear object boundary at high-resolution with limited GPU memory. It is worth noting that by introducing uncertainty into the training process, our HRRN can well address the high-resolution refinement task without using any high-resolution training data. 
Extensive experiments on high-resolution saliency datasets as well as some widely used saliency benchmarks show that the proposed method achieves superior performance compared to the state-of-the-art methods.
\end{abstract}

%%%%%%%%% BODY TEXT
\section{Introduction}
Salient object detection (SOD) is derived with the goal of accurately detecting and segmenting the most distinctive objects from visual scenes. As a preliminary step, it plays an essential role in various visual systems, such as
video object segmentation~\cite{DBLP:journals/pami/WangSYP18}, light field image segmentation~\cite{DBLP:conf/iccv/0002PLLZ19}, image-sentence matching~\cite{DBLP:conf/iccv/JiWHP19}, person re-identiﬁcation~\cite{DBLP:conf/ijcai/LiuZZJ20} and instance segmentation~\cite{DBLP:conf/cvpr/ZhouWJDY20}. 

Recently, the rapid development of the commodity imaging and display device, has resulted in higher requirements for the producing and editing of high-resolution (e.g., 720p, 1080p and 4K) images. Salient object detection as well as many state-of-the-art computer vision tasks are facing various challenges when encountering high-resolution scenarios. A good high-resolution salient object detection method should not only accurately detect the whole salient object but also predict the precise boundaries of salient objects. Despite the conventional Deep Neural Networks (DNNS) based SOD models have achieved remarkable performance at low-resolution (e.g., typical size $224 \times 224$, $384 \times 384$), they often fail to generate high quality detection results for high-resolution images. The major reason for this drawback is that the most previous methods try to identify the salient regions and estimate the accurate objects boundaries simultaneously in one step, which are two difficult and inherently different problems for high-resolution salient object detection. To address the first problem, a network is required to capture sufficient semantics by maintaining a larger receptive field. 
However, since the memory usage increases dramatically along with the image resolution, it is impractical for these models to directly learn sufficient semantics for high-resolution images.
One plausible way is introducing downsample operations, but the structure details are inevitably lost during the downsampling, which however is precisely the key to solving the second problem.

\begin{figure}
\centering
\includegraphics[scale=1,width = 8.5cm]{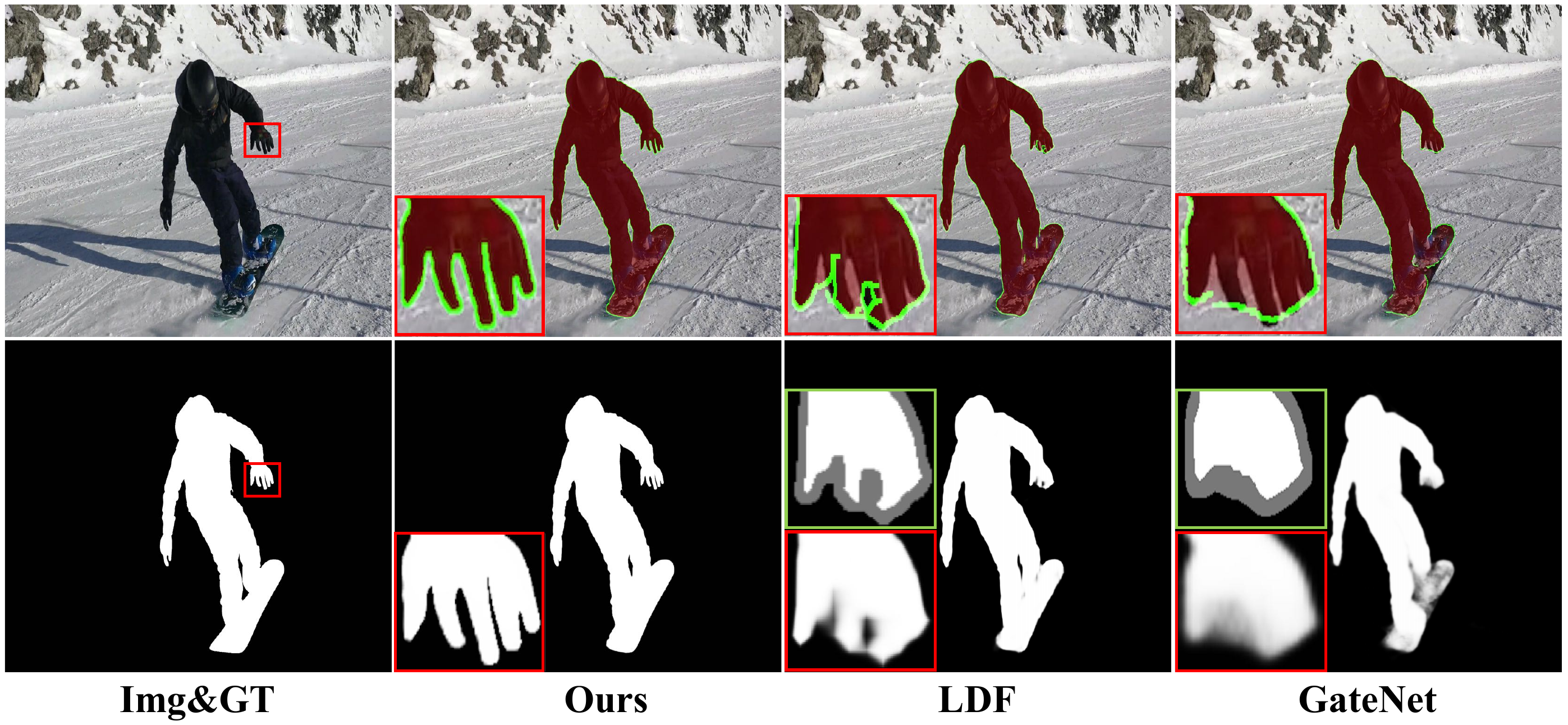}
\caption{Comparison with state-of-the-art method in high-resolution SOD. Best viewed by zooming in. }
\label{main-fig1}
\vspace{-0.5cm}
\end{figure}

Unfortunately, most of the existing low-resolution SOD methods~\cite{DBLP:conf/cvpr/LiuH16,DBLP:conf/iccv/ZhangWLWR17,DBLP:conf/eccv/ChenTWH18,DBLP:conf/cvpr/WeiWWSH020} try to address the aforementioned two problems with a single regression framework, which ignore the inherent difference between the two problems and result in blurry boundaries. As shown in Fig.\ref{main-fig1}, if we take a deeper look at the saliency map generated by the representative existing methods LDF~\cite{DBLP:conf/cvpr/WeiWWSH020} and GateNet~\cite{DBLP:journals/corr/abs-2007-08074}, we can observe that pixels can be divided into three different sets: (1) most pixels inside the salient object have the highest saliency value, and we call these pixels as \textit{definite salient pixels}; (2) most pixels in the background regions have the lowest salient value, which belong to \textit{definite background pixels}; (3) saliency values of the pixels at blurry object boundaries fluctuate between 0 and 1, so we call these pixels as \textit{uncertain pixels}. An ideal SOD method should effectively identify the definite salient and background regions in the image and accurately calculate the saliency value of pixels in the uncertain region to preserve a clear object boundary. From this perspective, there are essentially two tasks in SOD which demand quite different abilities to address the aforementioned two problems. The former task can be viewed as a classic classification task, while the later one is a typical regression task. 

Despite the demand for effective high-resolution SOD methods, this line of work is rarely studied. In this paper, motivated by the new observation that SOD should be disentangled into two tasks, we propose a novel deep learning framework for high-resolution salient object detection. Specifically, we decouple the high-resolution salient object detection into a low-resolution saliency classification network (LRSCN) and a high-resolution refinement network (HRRN). LRSCN is designed to capture sufficient semantics at low-resolution and classify the pixels into three different sets for later process. HRRN aims at accurately refining the saliency value of pixels in the uncertain region to preserve a clear object boundary at high-resolution with limited GPU memory. 
As discussed above, HRRN requires structure details in high-resolution image. However, widely used low-resolution saliency datasets generally have some problems in annotation quality~\cite{DBLP:conf/iccv/ZengZLZL19}, making it almost impossible to directly obtain enough object boundary details from these defective datasets to train the high-resolution network. 
In the very recent work, Zeng et al.~\cite{DBLP:conf/iccv/ZengZLZL19} proposed to train their SOD network by using high-resolution images with accurate annotation. However, such high-quality image annotation requires heavy labor costs. In our paper, we argue that it is unnecessary to use such accurately annotated high-resolution images in network training. By introducing uncertainty~\cite{DBLP:conf/nips/KendallG17} in the training process, our HRRN can well address the high-resolution refinement task only using the low-resolution training datasets with poor annotation. 

Our major contributions can be summarized as:

\begin{itemize}
\item We provide a new perspective that high-resolution salient object detection should be disentangled into two tasks, and demonstrate that the disentanglement of the two tasks is essential for improving the performance of DNN based SOD models.

\item Motivated by the principle of disentanglement, we propose a novel framework for high-resolution salient object detection, which uses LRSCN to capture sufficient semantics at low-resolution and HRRN for accurate boundary refinement at high-resolution.

\item We make the earliest efforts to introduce the uncertainty into SOD network training, which empowers HRRN to well address the high-resolution refinement task without any high-resolution training datasets.

\item We perform extensive experiments to demonstrate the proposed method refreshes the SOTA performance on high-resolution saliency datasets as well as some widely used saliency benchmarks by a large margin.
\end{itemize}

\begin{figure*}
\centering
\includegraphics[scale=0.52]{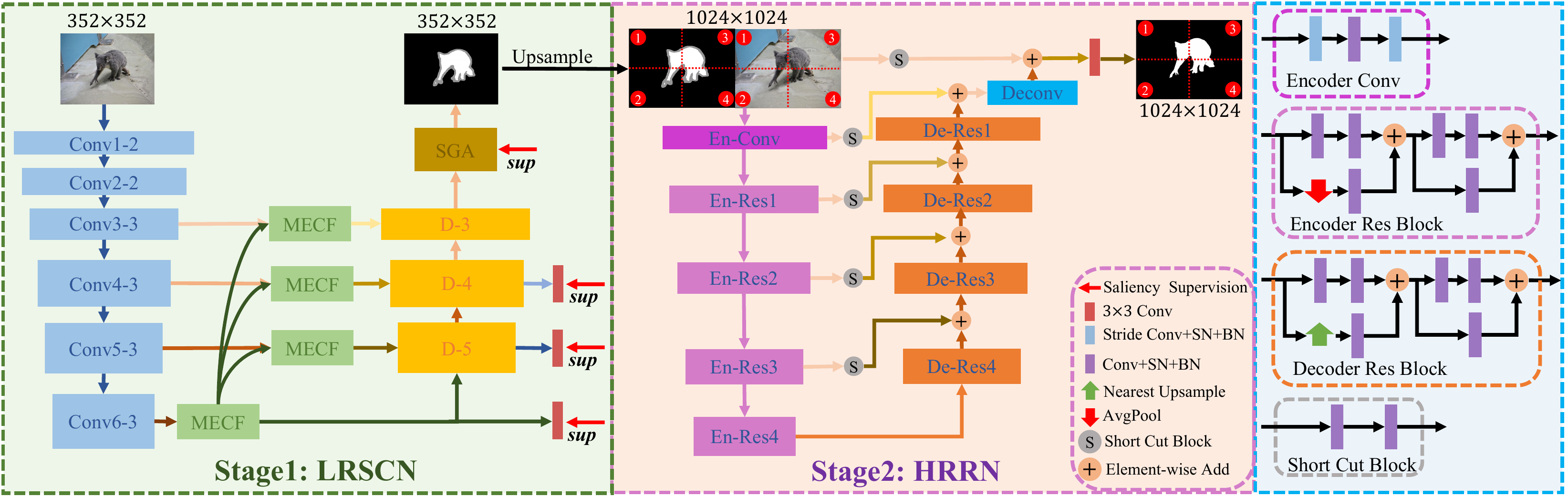}
\caption{The framework of the proposed disentangled high quality salient object detection method.}
\label{framework}
\vspace{-0.5cm}
\end{figure*}

\section{Related Work}
Over the past decades, a large amount of SOD algorithms have been developed. Traditional models~\cite{DBLP:journals/pami/IttiKN98,DBLP:journals/pami/ChengMHTH15,DBLP:journals/ijcv/WangJYCHZ17,DBLP:conf/eccv/WangZLSQ16,DBLP:conf/iccv/KleinF11}detect salient objects by utilizing various heuristic saliency priors with hand-crafted features. More details about the traditional methods can be found in the survey~\cite{DBLP:journals/cvm/BorjiCHJL19}. Recently, with the development of deep learning, the performance of saliency detection has archived great improvment~\cite{DBLP:conf/cvpr/LiuH16,DBLP:conf/eccv/FanCLGHB18,DBLP:conf/iccv/00380ZH19,DBLP:conf/aaai/SongYMFZMW20,DBLP:conf/aaai/ZhouWLWS20,8237695,DBLP:conf/cvpr/ZhangWQLW18}. Here we mainly focus on deep learning based saliency detection models.

Recently, some DNN-based models use various feature enhancement strategies to improve the ability of localization and awareness of salient objects~\cite{DBLP:journals/pami/HouCHBTT19,DBLP:conf/cvpr/WuSH19,DBLP:conf/aaai/WangCZZ0G20,DBLP:conf/aaai/ChenXCH20,DBLP:journals/corr/abs-1911-11445,DBLP:conf/cvpr/PangZZL20,DBLP:journals/corr/abs-2007-08074,DBLP:journals/corr/abs-2003-05643}, or take advantage of edge features to restore the structural details of salient objects~\cite{DBLP:conf/cvpr/WangZSHB19,DBLP:conf/iccv/ZhaoLFCYC19,DBLP:conf/iccv/WuSH19,DBLP:conf/cvpr/ZhouXLCY20}. For example, Pang et al.~\cite{DBLP:conf/cvpr/PangZZL20} applied the transformation-interaction-fusion strategy on multi-level and multi-scale features to learn discriminant feature representation. Zhao et al.~\cite{DBLP:journals/corr/abs-2007-08074} designed a gated dual branch structure to build the cooperation among different levels of features and improve the discriminability of the whole network.  In~\cite{DBLP:conf/iccv/ZhaoLFCYC19}, edge features from edge detection branch was fused with salient features as complementary information to enhance the structural details for accurate saliency detection. Zhou et al.~\cite{DBLP:conf/cvpr/ZhouXLCY20} 
used two individual branches for representing saliency and contour stream respectively, and a novel feature fusion module for their correlation combination.

Different from the above methods, some methods consider leveraging predict-refine architecture to generate fine salient objects. For example, Wang et al.~\cite{DBLP:conf/cvpr/WangZWL0RB18} proposed to localize salient objects globally and then refine them by a local boundary refinement module. Qin et al.~\cite{DBLP:conf/cvpr/QinZHGDJ19} was composed of an Encoder-Decoder network and a residual refinement module, which were respectively in charge of saliency prediction and saliency map refinement.

However, all these methods cannot handle high-resolution salient object detection problem well since such simple regression framework cannot identify the salient regions and estimate the accurate objects boundaries simultaneously and their architectures are not optimized for high-resolution SOD. Zeng et al.~\cite{DBLP:conf/iccv/ZengZLZL19} tried to alleviate this problem by leveraging both global semantic information and local high-resolution details to accurately detect salient objects in high-resolution images. However, Zeng et al.~\cite{DBLP:conf/iccv/ZengZLZL19} relies on high-resolution training images with accurate annotation, which requires heavy labor costs. Different from the above methods, we disentangle high-resolution SOD into two tasks at different resolutions: identifying the salient regions at low-resolution and estimating the accurate objects boundaries at high-resolution. Moreover, unlike Zeng et al.~\cite{DBLP:conf/iccv/ZengZLZL19}, we introduce novel uncertainty loss, which empowers our HRRN to well address the high-resolution refinement task without using any high-resolution training datasets. Recently, Wei et al.~\cite{DBLP:conf/cvpr/WeiWWSH020} and Zhang et al.~\cite{DBLP:journals/corr/abs-2007-12211} also 
leverage disentanglement in their SOD methods. However, they still try to address the SOD task under a single regression framework but with decoupled supervisions. Unlike our proposed methods, their disentanglement frameworks barely touch the very nature of the SOD, which essentially contains two different tasks. 
For more information about the DNN-based methods, please refer to survey~\cite{DBLP:journals/corr/abs-1904-09146,DBLP:journals/spm/HanZCLX18}.

\section{Proposed Method}
In this section, we first describe the overall architecture of the proposed disentangled high quality salient object detection network, then elaborate our main contributions, which are corresponding to LRSCN and HRRN.
\subsection{Network Overview}
The architecture of the proposed approach is illustrated in Fig.\ref{framework}. 
As can be seen, the disentanglement includes two decoupled tasks at two different resolutions.   
LRSCN aims at capturing sufficient semantics at low-resolution and classifying the pixels into three different sets, which also can save the memory usage. While  estimating the accurate objects boundaries needs more local details at high-resolution. So, we design HRRN to regress the saliency value of pixels and preserve a clear object boundary at high-resolution.

LRSCN has a simple U-Net like Encoder-Decoder architecture~\cite{DBLP:conf/miccai/RonnebergerFB15}. VGG-16~\cite{DBLP:journals/corr/SimonyanZ14a} is used as backbone. Following~\cite{DBLP:journals/pami/HouCHBTT19,DBLP:conf/iccv/ZhaoLFCYC19}, we connect another side path to the last pooling layer in VGG-16. Hence, we obtain six side features Conv1-2, Conv2-2, Conv3-3, Conv4-3, Conv5-3 and Conv6-3 from backbone network. Because Conv1-2 and Conv2-2 are too close to the input and their receptive fields are too small, following~\cite{DBLP:conf/iccv/ZhaoLFCYC19,DBLP:conf/cvpr/WuSH19}, we only use the last four levels features for the following process. Conv6-3 is denoted as $\{F_h|h = 6\}$, the other three levels features are denoted as $\{F_l|l = 3,4,5\}$. Multi-scale feature extraction and Cross-level feature fusion (MECF) module is added between encoder and decoder to help improve the discriminability of feature representations. Decoder fuses the output features from MECF and the upsampled features from the previous stage in a bottom-up manner. The output of each decoder is deﬁned as $\{D_i|i=3,4,5,6\}$. Finally, SGA module is built upon $D_3$ for accurate trimap $T$ generation. 

As described, LRSCN is classification task and aims at capturing sufficient semantics at low-resolution. To regress a clear object boundary value, the input of HRRN is a high-resolution image under guidance of the trimap provided by LRSCN. HRRN has a basic Encoder-Decoder architecture and with the help of uncertainty loss, the network can be more robust to noisy data and predict a high-resolution saliency map with clear boundary.

\subsection{Architecture of LRSCN}
To capture sufficient semantics at low-resolution, learning discriminant feature representations is essential. The network should not only consider scale and location variations of different salient objects, but also distinguish the appearance difference between the salient object and the non-salient regions. To achieve the first goal, we develop a multi-scale feature extraction module (ME) based on Global Convolutional Network (GCN)~\cite{DBLP:conf/cvpr/PengZYLS17} to enlarge the feature receptive field and obtain multi-scale information. To achieve the second goal, we utilize cross-level feature fusion module (CF) to leverage the advantages of features at different levels.Moreover, in designing the network architecture, inspired by~\cite{DBLP:conf/cvpr/XieGDTH17}, we use split-transform-merge strategy to further enlarge feature receptive fields and hence results in more discriminative feature representations. Specifically, we uniformly split the input $F$ into two portions $\{ F^1,F^2 \}$ by channel dimension, then $F^1$ is sent into multi-scale feature extraction pathway and $F^2$ is sent into cross-level feature fusion pathway. The outputs of these two pathways are concatenated together as the final output. we call this bridge module as \textbf{MECF module}, which is shown in Fig.\ref{MECF}. More details about MECF module can be found in section 6 of supplementary materials.

\begin{figure}
\centering
\includegraphics[scale=0.7,width = 8.5cm,height=3.8cm]{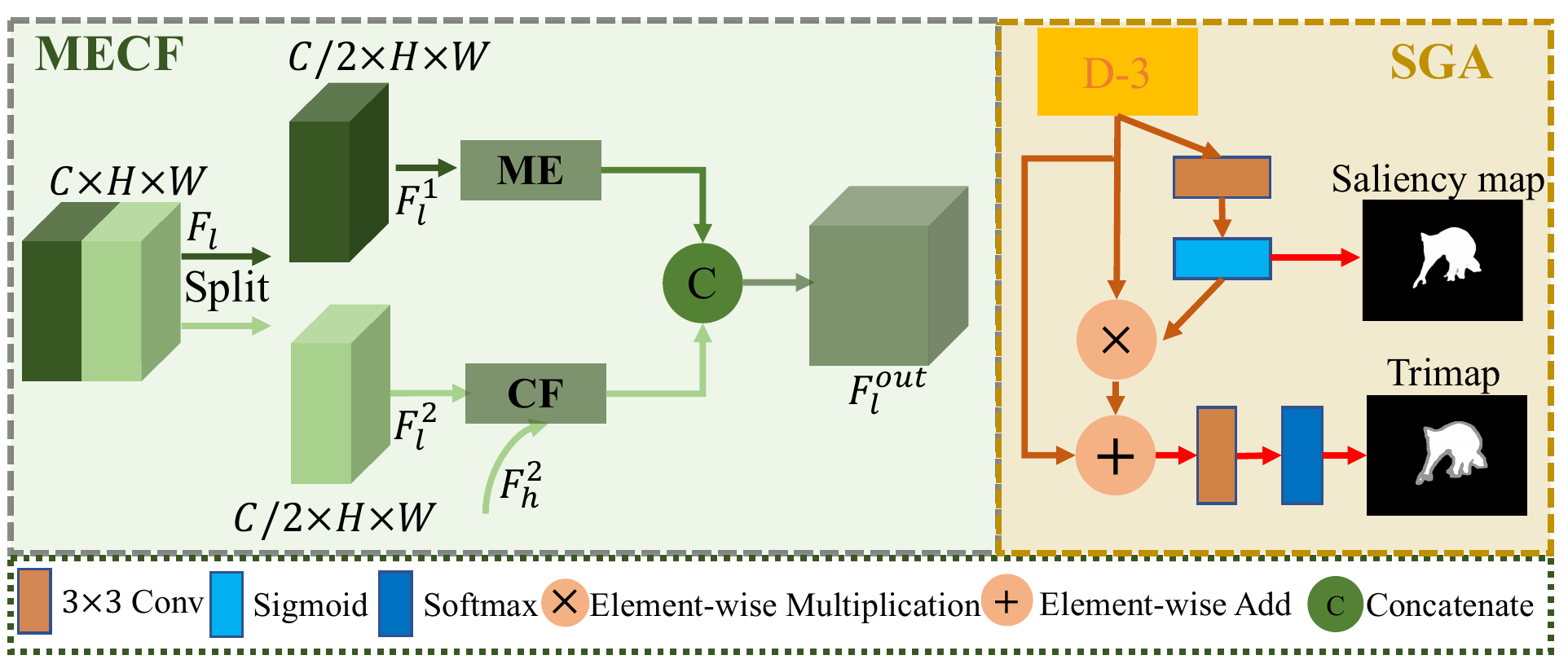}
\caption{Architecture of MECF and SGA Modules.}
\label{MECF}
\vspace{-0.5cm}
\end{figure}

\textbf{SGA Module.} As illustrated in Fig.\ref{framework}, each decoder fuses features from MECF module and previous decoder stage, then uses $3 \times 3$ convolutional layer for final prediction. To maintain consistency between trimap and saliency map and ensure the uncertain regions of the trimap can accurately cover the boundary of saliency map, we design a saliency guide attention module (SGA) on $D_3$. Specifically, we first use a $3 \times 3$ convolution and sigmoid function to compute a saliency map. Then, the saliency map is treated as spatial weight map which can help refine feature and generate an accurate trimap. 
Finally, the output trimap $T$ is 3-channel classiﬁcation logits. The whole SGA module guarantees the alignment of trimap and saliecny map.

\vspace{-0.2cm}
\subsection{Architecture of HRRN}
Following the principle of disentanglement, HRRN aims at accurately refining the saliency value of pixels in the uncertain region to preserve a clear object boundary at high-resolution under the guidance of the trimap provided by LRSCN. The architecture of HRRN is shown in Fig.\ref{framework}. HRRN has a simple U-NET like architecture. For better prediction at high-resolution, we then do some non-trivial modifications. First, lower-level features contain rich spatial and detail information which play a crucial role in restoring a clear object boundary, so decoder combines encoder features before each upsampling block instead of  after each upsampling block. Moreover, we use a two layers short cut block to align channels of encoder features for feature fusion. Second, to let network pay more attention to detail information, we directly feed the original input to the last convolutional layer through a short cut block to generate better results. Finally, learning from image generation tasks~\cite{DBLP:conf/iclr/BrockDS19,DBLP:conf/icml/ZhangGMO19}, we use the spectral normalization~\cite{DBLP:conf/iclr/MiyatoKKY18} to each convolutional layer to add a constraint on Lipschitz constant of the network and stable the training.

\vspace{-0.2cm}
\subsection{Loss Function of LRSCN}
\begin{figure}
\centering
\includegraphics[scale=0.4]{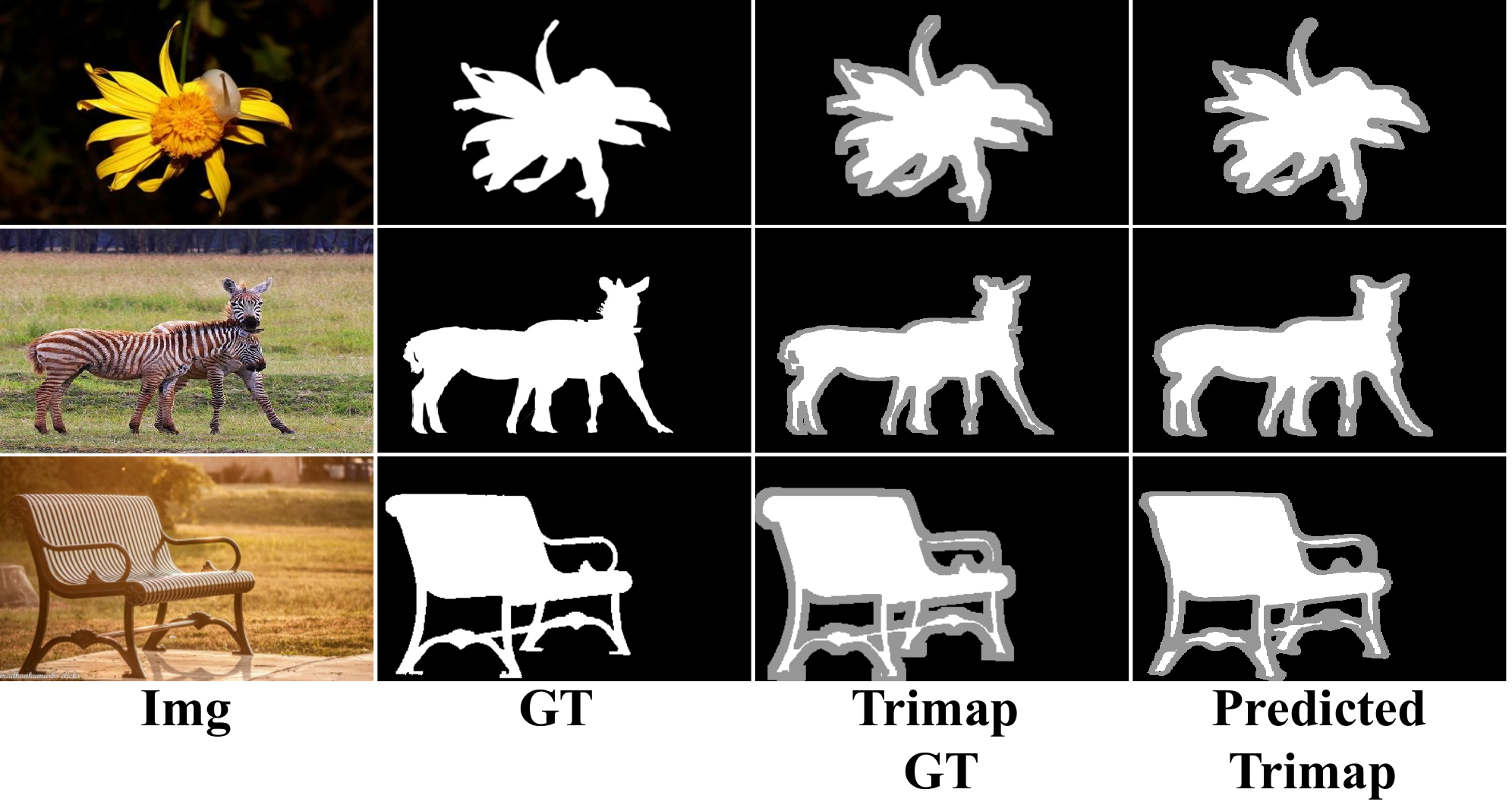}
\caption{Examples of trimap. Column 3 shows the trimaps generated from GT. Column 4 shows the trimaps predicted by LRSCN.}
\label{trimap}
\vspace{-0.5cm}
\end{figure}

To supervise LRSCN, inspired by some typical matting mehtods~\cite{DBLP:conf/iccv/CaiZFHLLLWS19,DBLP:conf/cvpr/TangAOGA19}, we generate trimap groundtruth $T^{gt}$, which can represent the definite salient, definite background and uncertain regions. As described, uncertain regions exist mainly at the boundaries of the objects. So we erase and dilate binary groundtruth maps at the object boundaries with a random pixel number (5,7,9,11,13) to generate the GT uncertain regions. The remaining foreground and background regions represent definite salient and background regions. $T^{gt}$ is defined as:
\begin{equation}
	T^{gt}(x,y) = 
	\begin{cases}
	2, \  &{ T^{gt}(x,y) \in definite \ salient} \\
	0, \  &{T^{gt}(x,y) \in definite \ background} \\
	1, \  &{T^{gt}(x,y) \in uncertain \  region}	
   \end{cases}
\end{equation}
where $(x,y)$ stands for each pixel location on the image. Some examples can be seen in Fig.\ref{trimap}.

For trimap supervision, we use Softmax cross-entropy loss, which is defined as:
\begin{equation}
\begin{aligned}
L_{trimap} = \frac{1}{N}\sum_{i} - log( \frac{e^{T_i}}{\sum_j e^{T_j}} ).
\end{aligned}
\end{equation}
To guarante the arruracy of trimap, we add extra saliency supervision $L_{saliency}$ as the supplement of trimap supervision. Similar to BASNet~\cite{DBLP:conf/cvpr/QinZHGDJ19}, we use pixel-level, region-level and object-level supervision strategy on multi-levels to better keep the uniformity and wholeness of the salient objects. Specifically, binary cross-entropy (BCE)~\cite{DBLP:journals/anor/BoerKMR05}, SSIM~\cite{DBLP:journals/tip/WangBSS04} and F-measure loss~\cite{DBLP:conf/iccv/ZhaoGWC19} are denoted as pixel-level, region-level and object-level loss. Note that all parts of LRSCN are trained jointly, so the overall loss function is given as:
\begin{equation}
L_{LRSCN} = L_{saliency} + L_{trimap}.
\end{equation}
We do not use uncertainty loss because the main goal of LRSCN is to capture sufficient semantics, not accurate boundary. More details about $L_{saliency}$ can be found in section 5 of supplementary materials.

\subsection{Loss Function of HRRN}
We perform a $L_1$ loss and novel uncertainty loss to restore the fine structures and boundaries of salient objects. For an input high-resolution image $I$, let $G^{H}$ denote its groundtruth, and predicted saliency map is $S^H$. 

We leverage the $L_1$ loss to compare an absolute difference between predicted saliency map and groundtruth over the definite salient and background regions:
\begin{equation}
L_1 = \frac{1}{E}\sum_{i \in E}|S^H_i - G^H_i|,
\end{equation}
where $E$ indicates the number of pixels which are labeled as definite salient or background in the trimap, $S^H_i$ and $G^H_i$ denote the predicted and groundtruth value at position $i$.

We cannot directly compute $L_1$ loss between predicted saliency map and groundtruth over the uncertain regions because widely used saliency training datasets have some problems in annotation quality~\cite{DBLP:conf/iccv/ZengZLZL19}. We show these low quality annotations in section 4 of supplementary materials. It is almost impossible to directly obtain enough object boundary details from these defective datasets to train the high-resolution network. To address this problem, we design uncertainty loss, which empowers our HRRN to well address the high-resolution refinement task only using these defective low-resolution training datasets. It is worth noting that there are some previous works ~\cite{DBLP:conf/cvpr/ZhangFDASZB20,DBLP:conf/iccv/ZhangWLWY17} involving ``uncertainty'' in their titles, which seem relevant to our method.
However, in~\cite{DBLP:conf/cvpr/ZhangFDASZB20}, ``uncertainty'' means the human perceptual uncertainty modeled by CVAE.  While in~\cite{DBLP:conf/iccv/ZhangWLWY17}, ``uncertainty'' indicates the saliency prediction system uncertainty modeled by R-dropout. Obviously, their usages of uncertainty are different from ours.

Inspired by~\cite{DBLP:conf/nips/KendallG17}, Gaussian likelihood is used to model the uncertainty. Let $x$ and $f(x)$ be the input and output of HRRN, and Gaussian likelihood is defined as:
\begin{equation}
p(y|f(x)) = \mathcal{N}(f(x), \sigma^2),
\end{equation}
where $\sigma$ measures uncertainty of the estimation, $y$ is the label of output. In maximum likelihood inference, we maximize the log likelihood of the model, which is written as:
\begin{equation}
log p(y|f(x)) \propto - \frac{||y - f(x)||^2}{2\sigma^2} - \frac{1}{2}log\sigma^2,
\end{equation}
so the proposed uncertainty loss is defined as:
\begin{equation}
L_{uncertainty} =  \frac{||y - f(x)||^2}{2\sigma^2} + \frac{1}{2}log\sigma^2.
\end{equation}
We only care about the pixels in uncertain regions, so $L_{uncertainty}$ is written as:
\begin{equation}
L_{uncertainty} = \frac{1}{U}\sum_{i \in U}\frac{||S^H_i - G^H_i||^2}{2\sigma_{i}^2} + \frac{1}{2}log\sigma_{i}^2,
\end{equation}
where $U$ is the total number of pixels in uncertain region, $\sigma_{i}$ is the uncertainty of each pixel and is generated from HRRN. Different from directly learning from noisy data, uncertainty loss can allow the network to learn how to attenuate the effect from erroneous labels. Specifically, pixels for which the network learned to predict high uncertainty will have a smaller value of the first term of Eq.8, so have little effect on the loss. Meanwhile, large uncertainty increases the contribution of the second term of Eq.8, and in turn penalizes the model and lets the model make a better prediction that has low uncertainty.
Note that all parts of HRRN are trained jointly, so the over all loss function is given as:
\begin{equation}
L_{HRRN} = L_{uncertainty} + L_{1}.
\end{equation}

\begin{figure}
\centering
\includegraphics[scale=0.32]{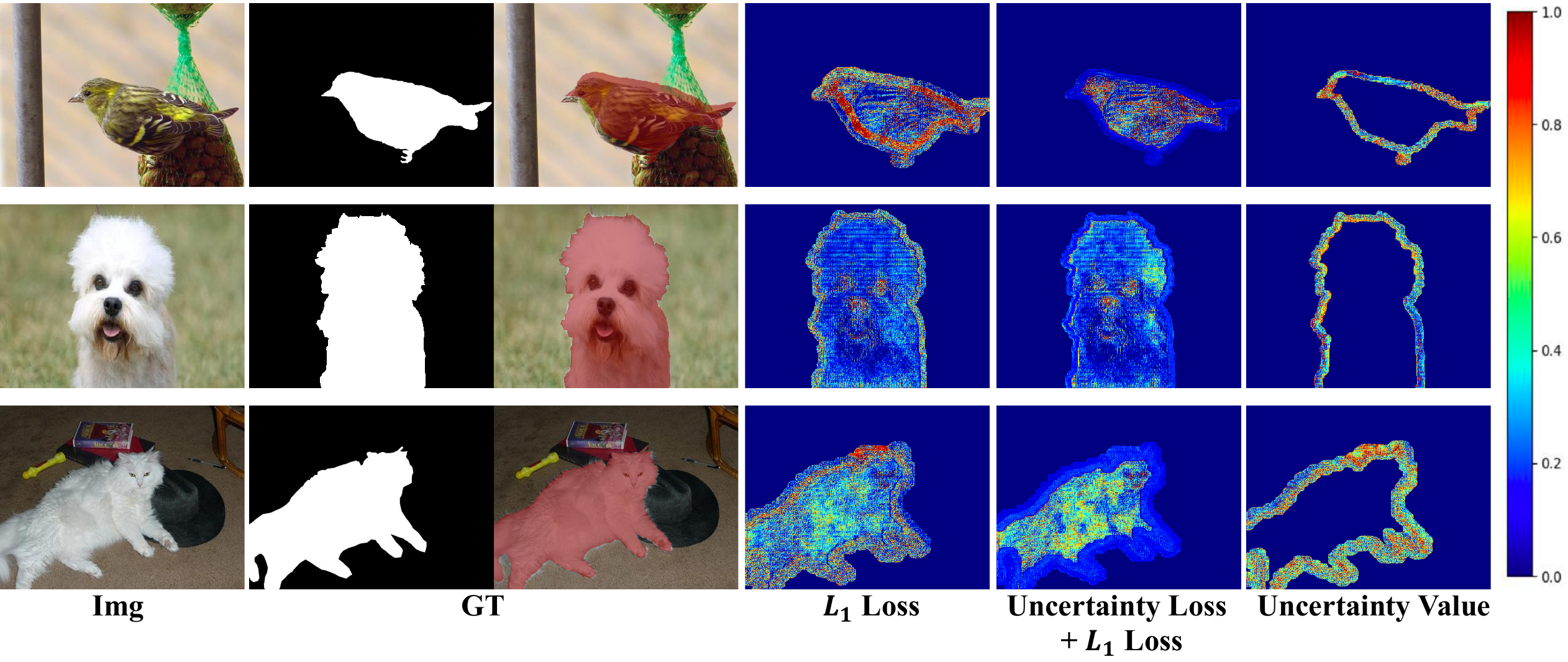}
\caption{The impact of the losses. Best viewed by zooming in.}
\label{uncertainty_loss}
\vspace{-0.5cm}
\end{figure}

To show how proposed uncertainty loss makes the network attenuate the effect from erroneous labels during training, we visualize the impact of $L_1$ loss and uncertainty loss at the same training iteration in Fig.\ref{uncertainty_loss}. We show the impact of both these losses in the same image. The images shown in Fig.\ref{uncertainty_loss} are these which have problems in annotation quality. Compared with column.4 and column.5, if we only use $L_1$ loss, the weight of the loss in the uncertainty region will be large, which leads the network hard to converge. While uncertainty loss will make the weight of the loss in the uncertainty region be small and let the network ignore effects from noisy data as much as possible. Column.6 shows the uncertainty value of pixels in uncertain regions. It can be seen that pixels in uncertain regions usually have higher uncertainty value. In general, compared to $L_1$ loss, uncertainty loss reduces the weight of the loss in the uncertainty region, thus mitigating the impact of noisy data on the network. But due to the uncertainty value, it will allow the network to learn how to predict a better prediction with a low certainty value, instead of ignoring the learning of the uncertainty region completely. These visual comparisons show how uncertainty loss makes network more robust to noisy data.

\begin{table*}
\caption{Quantitative comparison with SOTA on two high-resolution and three low-resolution datasets. The best three results are in {\color[HTML]{FF0000} red} , {\color[HTML]{32CB00} green} and {\color[HTML]{3166FF} blue} fonts. "$\dagger$" means the results are post-processed by dense conditional random field(CRF)~\cite{DBLP:conf/nips/KrahenbuhlK11}. "*" means using ResNeXt-101~\cite{DBLP:conf/cvpr/XieGDTH17} backbone. "$\star$" means using ResNet-101 backbone. "$\ddag$" means using Res2Net50~\cite{gao2019res2net} backbone. MK: MSRA10K~\cite{DBLP:journals/pami/ChengMHTH15}, DUTS: DUTS-TR~\cite{DBLP:conf/cvpr/WangLWF0YR17}, MB: MSRA-B~\cite{DBLP:journals/pami/LiuYSWZTS11}, HR: HRSOD-Training~\cite{DBLP:conf/iccv/ZengZLZL19}, HR-L: HRSOD-Training resized in low-resolution. Smaller MAE, BDE and $B_\mu$, larger $F_{\beta}^{max}$, $F_{\beta}$ and $S_m$ correspond to better performance. }
\centering
\scalebox{0.45}{
\begin{tabular}{cccccccccccccccccccccccccc}
\hline
\multicolumn{1}{c|}{}                         & \multicolumn{1}{c|}{}                                                                              & \multicolumn{6}{c|}{HRSOD-TE}                                                                                                                                                                                 & \multicolumn{6}{c|}{DAVIS-S}                                                                                                                                                                                  & \multicolumn{4}{c|}{DUT-OMRON}                                                                                                                                     & \multicolumn{4}{c|}{DUTS-TE}                                                                                                                   & \multicolumn{4}{c}{HKU-IS}                                                                                                \\ \cline{3-26} 
\multicolumn{1}{c|}{\multirow{-2}{*}{Models}} & \multicolumn{1}{c|}{\multirow{-2}{*}{\begin{tabular}[c]{@{}c@{}}Training\\ datasets\end{tabular}}} & $F_{\beta}^{max}$             & $F_\beta$                    & $S_m$                        & MAE                          & BDE                           & \multicolumn{1}{c|}{$B_\mu$}                      & $F_\beta^{max}$              & $F_\beta$                    & $S_m$                        & MAE                          & BDE                           & \multicolumn{1}{c|}{$B_\mu$}                      & $F_\beta^{max}$                                  & $F_\beta$                    & $S_m$                        & \multicolumn{1}{c|}{MAE}                          & $F_\beta^{max}$              & $F_\beta$                    & $S_m$                        & \multicolumn{1}{c|}{MAE}                          & $F_\beta^{max}$              & $F_\beta$                    & $S_m$                        & MAE                          \\ \hline
\multicolumn{26}{c}{VGG-16 backbone}                                                                                                                                                                                                                                                                                                                                                                                                                                                                                                                                                                                                                                                                                                                                                                                                                                                                                                                                                                                                 \\ \hline
\multicolumn{1}{c|}{Amulet(ICCV2017)}         & \multicolumn{1}{c|}{MK}                                                                            & 0.799                        & 0.717                        & 0.829                        & 0.075                        & 139.889                       & \multicolumn{1}{c|}{0.947}                        & 0.802                        & 0.755                        & 0.848                        & 0.042                        & 64.827                        & \multicolumn{1}{c|}{0.856}                        & 0.743                                            & 0.647                        & 0.781                        & \multicolumn{1}{c|}{0.098}                        & 0.778                        & 0.678                        & 0.804                        & \multicolumn{1}{c|}{0.085}                        & 0.897                        & 0.841                        & 0.886                        & 0.051                        \\
\multicolumn{1}{c|}{DGRL(CVPR2018)}           & \multicolumn{1}{c|}{DUTS}                                                                          & 0.821                        & 0.789                        & 0.847                        & 0.055                        & 95.034                        & \multicolumn{1}{c|}{0.889}                        & 0.803                        & 0.772                        & 0.859                        & 0.038                        & 50.323                        & \multicolumn{1}{c|}{0.826}                        & 0.774                                            & 0.709                        & 0.810                        & \multicolumn{1}{l|}{0.063}                        & 0.828                        & 0.794                        & 0.842                        & \multicolumn{1}{c|}{0.050}                        & 0.910                        & \multicolumn{1}{l}{0.881}    & \multicolumn{1}{l}{0.896}    & \multicolumn{1}{l}{0.037}    \\
\multicolumn{1}{c|}{DSS$\dagger$(TPAMI2019)}           & \multicolumn{1}{c|}{MB}                                                                            & 0.826                        & 0.756                        & 0.840                        & 0.060                        & 145.403                       & \multicolumn{1}{c|}{0.952}                        & 0.830                        & 0.728                        & 0.865                        & 0.041                        & 94.069                        & \multicolumn{1}{c|}{0.890}                        & 0.781                                            & 0.740                        & 0.790                        & \multicolumn{1}{c|}{0.062}                        & \multicolumn{1}{l}{0.825}    & 0.808                        & 0.820                        & \multicolumn{1}{c|}{0.057}                        & 0.916                        & 0.902                        & 0.878                        & 0.040                        \\
\multicolumn{1}{c|}{CPD(CVPR2019)}            & \multicolumn{1}{c|}{DUTS}                                                                          & 0.876                        & 0.829                        & 0.887                        & 0.039                        & {\color[HTML]{0070C0} 72.686} & \multicolumn{1}{c|}{{\color[HTML]{0070C0} 0.824}} & 0.878                        & 0.822                        & 0.903                        & 0.025                        & 36.649                        & \multicolumn{1}{c|}{{\color[HTML]{0070C0} 0.703}} & 0.794                                            & {\color[HTML]{0070C0} 0.745} & 0.818                        & \multicolumn{1}{c|}{{\color[HTML]{0070C0} 0.057}} & 0.864                        & 0.813                        & 0.867                        & \multicolumn{1}{c|}{0.043}                        & 0.924                        & 0.896                        & 0.904                        & {\color[HTML]{0070C0} 0.033} \\
\multicolumn{1}{c|}{EGNet(ICCV2019)}          & \multicolumn{1}{c|}{DUTS}                                                                          & 0.883                        & 0.814                        & 0.888                        & 0.044                        & 73.500                        & \multicolumn{1}{c|}{0.896}                        & 0.886                        & 0.794                        & 0.897                        & 0.030                        & 37.369                        & \multicolumn{1}{c|}{0.799}                        & {\color[HTML]{00B050} 0.803}                     & 0.744                        & 0.813                        & \multicolumn{1}{c|}{{\color[HTML]{0070C0} 0.057}} & {\color[HTML]{0070C0} 0.877} & 0.800                        & 0.866                        & \multicolumn{1}{c|}{0.044}                        & 0.927                        & 0.893                        & {\color[HTML]{0070C0} 0.910} & 0.035                        \\
\multicolumn{1}{c|}{MINet(CVPR2020)}          & \multicolumn{1}{c|}{DUTS}                                                                          & 0.902                        & 0.851                        & 0.903                        & 0.032                        & 76.291                        & \multicolumn{1}{c|}{0.849}                        & {\color[HTML]{0070C0} 0.915} & 0.864                        & {\color[HTML]{0070C0} 0.926} & {\color[HTML]{0070C0} 0.019} & {\color[HTML]{0070C0} 32.304} & \multicolumn{1}{c|}{0.742}                        & 0.794                                            & 0.741                        & {\color[HTML]{0070C0} 0.822} & \multicolumn{1}{c|}{{\color[HTML]{0070C0} 0.057}} & {\color[HTML]{0070C0} 0.877} & {\color[HTML]{0070C0} 0.823} & {\color[HTML]{0070C0} 0.875} & \multicolumn{1}{c|}{{\color[HTML]{0070C0} 0.039}} & {\color[HTML]{0070C0} 0.930} & {\color[HTML]{0070C0} 0.904} & {\color[HTML]{00B050} 0.912} & {\color[HTML]{00B050} 0.031} \\
\multicolumn{1}{c|}{ITSD(CVPR2020)}           & \multicolumn{1}{c|}{DUTS}                                                                          & 0.824                        & 0.715                        & 0.834                        & 0.071                        & 139.943                       & \multicolumn{1}{c|}{0.924}                        & 0.806                        & 0.687                        & 0.843                        & 0.055                        & 92.864                        & \multicolumn{1}{c|}{0.861}                        & {\color[HTML]{0070C0} 0.802}                     & {\color[HTML]{0070C0} 0.745} & {\color[HTML]{00B050} 0.828} & \multicolumn{1}{c|}{0.063}                        & 0.876                        & 0.798                        & {\color[HTML]{00B050} 0.877} & \multicolumn{1}{c|}{0.042}                        & 0.927                        & 0.890                        & 0.906                        & 0.035                        \\
\multicolumn{1}{c|}{GateNet(ECCV2020)}        & \multicolumn{1}{c|}{DUTS}                                                                          & {\color[HTML]{0070C0} 0.905} & 0.825                        & {\color[HTML]{0070C0} 0.906} & 0.035                        & 79.468                        & \multicolumn{1}{c|}{0.886}                        & 0.914                        & 0.825                        & 0.923                        & 0.023                        & 44.827                        & \multicolumn{1}{c|}{0.778}                        & 0.794                                            & 0.723                        & 0.821                        & \multicolumn{1}{c|}{0.061}                        & 0.870                        & 0.783                        & 0.870                        & \multicolumn{1}{c|}{0.045}                        & 0.929                        & 0.889                        & {\color[HTML]{0070C0} 0.910} & 0.036                        \\
\multicolumn{1}{c|}{HRNet(ICCV2019)}          & \multicolumn{1}{c|}{DUTS+HR}                                                                       & {\color[HTML]{0070C0} 0.905} & {\color[HTML]{0070C0} 0.888} & 0.897                        & {\color[HTML]{0070C0} 0.030} & 88.017                        & \multicolumn{1}{c|}{0.888}                        & 0.899                        & {\color[HTML]{0070C0} 0.888} & 0.876                        & 0.026                        & 44.359                        & \multicolumn{1}{c|}{0.801}                        & 0.743                                            & 0.690                        & 0.762                        & \multicolumn{1}{c|}{0.065}                        & 0.835                        & 0.788                        & 0.824                        & \multicolumn{1}{c|}{0.050}                        & 0.910                        & 0.886                        & 0.877                        & 0.042                        \\
\multicolumn{1}{c|}{Ours}                     & \multicolumn{1}{c|}{DUTS}                                                                          & {\color[HTML]{00B050} 0.918} & {\color[HTML]{00B050} 0.902} & {\color[HTML]{00B050} 0.912} & {\color[HTML]{00B050} 0.027} & {\color[HTML]{00B050} 48.468} & \multicolumn{1}{c|}{{\color[HTML]{00B050} 0.711}} & {\color[HTML]{00B050} 0.933} & {\color[HTML]{00B050} 0.919} & {\color[HTML]{00B050} 0.933} & {\color[HTML]{00B050} 0.015} & {\color[HTML]{00B050} 15.676} & \multicolumn{1}{c|}{{\color[HTML]{00B050} 0.536}} & \multicolumn{1}{l}{{\color[HTML]{FF0000} 0.804}} & {\color[HTML]{FF0000} 0.769} & {\color[HTML]{FF0000} 0.829} & \multicolumn{1}{c|}{{\color[HTML]{00B050} 0.053}} & {\color[HTML]{00B050} 0.882} & {\color[HTML]{00B050} 0.855} & {\color[HTML]{FF0000} 0.879} & \multicolumn{1}{c|}{{\color[HTML]{00B050} 0.036}} & {\color[HTML]{FF0000} 0.935} & {\color[HTML]{FF0000} 0.918} & {\color[HTML]{FF0000} 0.913} & {\color[HTML]{FF0000} 0.029} \\
\multicolumn{1}{c|}{Ours-DH}                  & \multicolumn{1}{c|}{DUTS+HR-L}                                                                     & {\color[HTML]{FF0000} 0.921} & {\color[HTML]{FF0000} 0.907} & {\color[HTML]{FF0000} 0.917} & {\color[HTML]{FF0000} 0.024} & {\color[HTML]{FF0000} 45.462} & \multicolumn{1}{c|}{{\color[HTML]{FF0000} 0.706}} & {\color[HTML]{FF0000} 0.938} & {\color[HTML]{FF0000} 0.926} & {\color[HTML]{FF0000} 0.936} & {\color[HTML]{FF0000} 0.014} & {\color[HTML]{FF0000} 14.412} & \multicolumn{1}{c|}{{\color[HTML]{FF0000} 0.531}} & 0.795                                            & {\color[HTML]{00B050} 0.764} & 0.820                        & \multicolumn{1}{c|}{{\color[HTML]{FF0000} 0.052}} & {\color[HTML]{FF0000} 0.894} & {\color[HTML]{FF0000} 0.865} & {\color[HTML]{FF0000} 0.879} & \multicolumn{1}{c|}{{\color[HTML]{FF0000} 0.035}} & {\color[HTML]{00B050} 0.933} & {\color[HTML]{00B050} 0.914} & 0.904                        & {\color[HTML]{00B050} 0.031} \\ \hline
\multicolumn{26}{c}{ResNet-50/ResNet-101/ResNeXt-101/Res2Net50 backbone}                                                                                                                                                                                                                                                                                                                                                                                                                                                                                                                                                                                                                                                                                                                                                                                                                                                                                                                                                             \\ \hline
\multicolumn{1}{c|}{R3Net*(IJCAI2018)}         & \multicolumn{1}{c|}{MK}                                                                            & 0.798                        & 0.744                        & 0.812                        & 0.081                        & 108.910                       & \multicolumn{1}{c|}{0.931}                        & 0.806                        & 0.753                        & 0.835                        & 0.041                        & 47.373                        & \multicolumn{1}{c|}{0.868}                        & 0.785                                            & 0.690                        & 0.819                        & \multicolumn{1}{c|}{0.073}                        & 0.778                        & 0.716                        & 0.837                        & \multicolumn{1}{c|}{0.067}                        & 0.915                        & 0.853                        & 0.894                        & 0.047                        \\
\multicolumn{1}{c|}{BASNet(CVPR2019)}         & \multicolumn{1}{c|}{DUTS}                                                                          & 0.878                        & 0.831                        & 0.890                        & 0.038                        & 67.643                        & \multicolumn{1}{c|}{0.823}                        & 0.857                        & 0.806                        & 0.881                        & 0.039                        & 46.283                        & \multicolumn{1}{c|}{{\color[HTML]{0070C0} 0.705}} & 0.805                                            & 0.766                        & 0.838                        & \multicolumn{1}{c|}{0.056}                        & 0.859                        & 0.791                        & 0.866                        & \multicolumn{1}{c|}{0.048}                        & 0.928                        & 0.895                        & 0.909                        & 0.032                        \\
\multicolumn{1}{c|}{PFPN$\star$(AAAI2020)}           & \multicolumn{1}{c|}{DUTS}                                                                          & 0.889                        & 0.825                        & 0.897                        & 0.042                        & 65.048                        & \multicolumn{1}{c|}{0.896}                        & 0.886                        & 0.822                        & 0.912                        & 0.025                        & 30.488                        & \multicolumn{1}{c|}{0.848}                        & {\color[HTML]{00B050} 0.818}                     & 0.748                        & {\color[HTML]{0070C0} 0.841} & \multicolumn{1}{c|}{0.057}                        & 0.885                        & 0.805                        & 0.887                        & \multicolumn{1}{c|}{0.041}                        & 0.937                        & 0.896                        & 0.919                        & 0.033                        \\
\multicolumn{1}{c|}{GCPA(AAAI2020)}           & \multicolumn{1}{c|}{DUTS}                                                                          & 0.889                        & 0.827                        & 0.894                        & 0.039                        & 70.320                        & \multicolumn{1}{c|}{0.873}                        & 0.912                        & 0.833                        & {\color[HTML]{0070C0} 0.924} & 0.021                        & {\color[HTML]{0070C0} 24.132} & \multicolumn{1}{c|}{0.759}                        & 0.812                                            & 0.748                        & 0.838                        & \multicolumn{1}{c|}{0.056}                        & 0.888                        & 0.817                        & {\color[HTML]{00B050} 0.891} & \multicolumn{1}{c|}{0.038}                        & 0.938                        & 0.898                        & 0.920                        & 0.031                        \\
\multicolumn{1}{c|}{F3N(AAAI2020)}            & \multicolumn{1}{c|}{DUTS}                                                                          & 0.900                        & 0.853                        & 0.897                        & 0.035                        & 65.901                        & \multicolumn{1}{c|}{0.817}                        & {\color[HTML]{0070C0} 0.915} & 0.845                        & 0.913                        & 0.020                        & 45.106                        & \multicolumn{1}{c|}{0.719}                        & 0.813                                            & 0.766                        & 0.838                        & \multicolumn{1}{c|}{0.053}                        & 0.891                        & 0.840                        & 0.888                        & \multicolumn{1}{c|}{0.035}                        & 0.937                        & 0.910                        & 0.917                        & {\color[HTML]{0070C0} 0.028} \\
\multicolumn{1}{c|}{LDF(CVPR2020)}            & \multicolumn{1}{c|}{DUTS}                                                                          & {\color[HTML]{0070C0} 0.905} & {\color[HTML]{0070C0} 0.866} & {\color[HTML]{0070C0} 0.905} & {\color[HTML]{0070C0} 0.032} & {\color[HTML]{0070C0} 58.655} & \multicolumn{1}{c|}{{\color[HTML]{0070C0} 0.812}} & 0.911                        & {\color[HTML]{0070C0} 0.864} & 0.922                        & {\color[HTML]{0070C0} 0.019} & 35.496                        & \multicolumn{1}{c|}{0.713}                        & {\color[HTML]{0070C0} 0.817}                     & {\color[HTML]{0070C0} 0.773} & 0.839                        & \multicolumn{1}{c|}{{\color[HTML]{0070C0} 0.052}} & {\color[HTML]{0070C0} 0.894} & {\color[HTML]{0070C0} 0.855} & {\color[HTML]{0070C0} 0.890} & \multicolumn{1}{c|}{{\color[HTML]{0070C0} 0.034}} & {\color[HTML]{0070C0} 0.939} & {\color[HTML]{0070C0} 0.914} & 0.919                        & {\color[HTML]{0070C0} 0.028} \\
\multicolumn{1}{c|}{CSF$\ddag$(ECCV2020)}          & \multicolumn{1}{c|}{DUTS}                                                                          & 0.894                        & 0.832                        & 0.900                        & 0.038                        & 71.293                        & \multicolumn{1}{c|}{0.922}                        & 0.899                        & 0.822                        & 0.912                        & 0.025                        & 30.488                        & \multicolumn{1}{c|}{0.848}                        & 0.815                                            & 0.750                        & 0.838                        & \multicolumn{1}{c|}{0.055}                        & {\color[HTML]{0070C0} 0.894} & 0.823                        & {\color[HTML]{0070C0} 0.890} & \multicolumn{1}{c|}{0.038}                        & 0.935                        & 0.902                        & {\color[HTML]{0070C0} 0.921} & 0.030                        \\
\multicolumn{1}{c|}{Ours}                     & \multicolumn{1}{c|}{DUTS}                                                                          & {\color[HTML]{00B050} 0.915} & {\color[HTML]{00B050} 0.902} & {\color[HTML]{00B050} 0.919} & {\color[HTML]{00B050} 0.024} & {\color[HTML]{00B050} 47.804} & \multicolumn{1}{c|}{{\color[HTML]{00B050} 0.750}} & {\color[HTML]{00B050} 0.935} & {\color[HTML]{00B050} 0.923} & {\color[HTML]{00B050} 0.937} & {\color[HTML]{00B050} 0.013} & {\color[HTML]{00B050} 14.396} & \multicolumn{1}{c|}{{\color[HTML]{00B050} 0.576}} & {\color[HTML]{00B050} 0.818}                     & {\color[HTML]{00B050} 0.785} & {\color[HTML]{00B050} 0.842} & \multicolumn{1}{c|}{{\color[HTML]{00B050} 0.051}} & {\color[HTML]{00B050} 0.895} & {\color[HTML]{00B050} 0.870} & {\color[HTML]{FF0000} 0.892} & \multicolumn{1}{c|}{{\color[HTML]{00B050} 0.033}} & {\color[HTML]{00B050} 0.943} & {\color[HTML]{00B050} 0.928} & {\color[HTML]{FF0000} 0.923} & {\color[HTML]{FF0000} 0.025} \\
\multicolumn{1}{c|}{Ours-DH}                  & \multicolumn{1}{c|}{DUTS}                                                                          & {\color[HTML]{FF0000} 0.922} & {\color[HTML]{FF0000} 0.909} & {\color[HTML]{FF0000} 0.922} & {\color[HTML]{FF0000} 0.022} & {\color[HTML]{FF0000} 46.495} & \multicolumn{1}{c|}{{\color[HTML]{FF0000} 0.746}} & {\color[HTML]{FF0000} 0.938} & {\color[HTML]{FF0000} 0.926} & {\color[HTML]{FF0000} 0.939} & {\color[HTML]{FF0000} 0.012} & {\color[HTML]{FF0000} 14.266} & \multicolumn{1}{c|}{{\color[HTML]{FF0000} 0.571}} & {\color[HTML]{FF0000} 0.820}                     & {\color[HTML]{FF0000} 0.791} & {\color[HTML]{FF0000} 0.843} & \multicolumn{1}{c|}{{\color[HTML]{FF0000} 0.048}} & {\color[HTML]{FF0000} 0.900} & {\color[HTML]{FF0000} 0.876} & {\color[HTML]{FF0000} 0.892} & \multicolumn{1}{c|}{{\color[HTML]{FF0000} 0.031}} & {\color[HTML]{FF0000} 0.944} & {\color[HTML]{FF0000} 0.929} & {\color[HTML]{00B050} 0.922} & {\color[HTML]{00B050} 0.026} \\ \hline
\end{tabular}}
\label{total}
\end{table*}

\begin{figure*}[!hbt]
\centering
\includegraphics[scale=0.5]{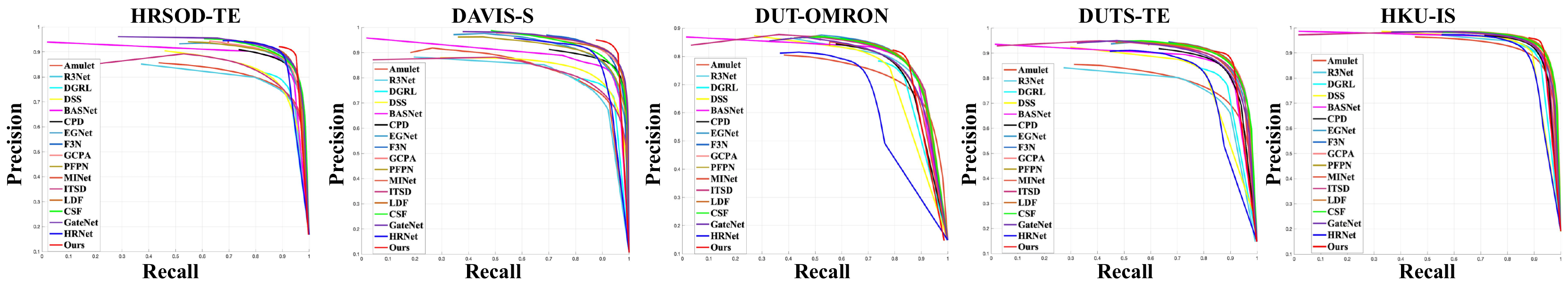}
\caption{Comparison of PR curves across two high-resolution and three low-resolution datasets.}
\label{PR}
\end{figure*}

\begin{figure*}[!hbt]
\centering
\includegraphics[scale=0.5]{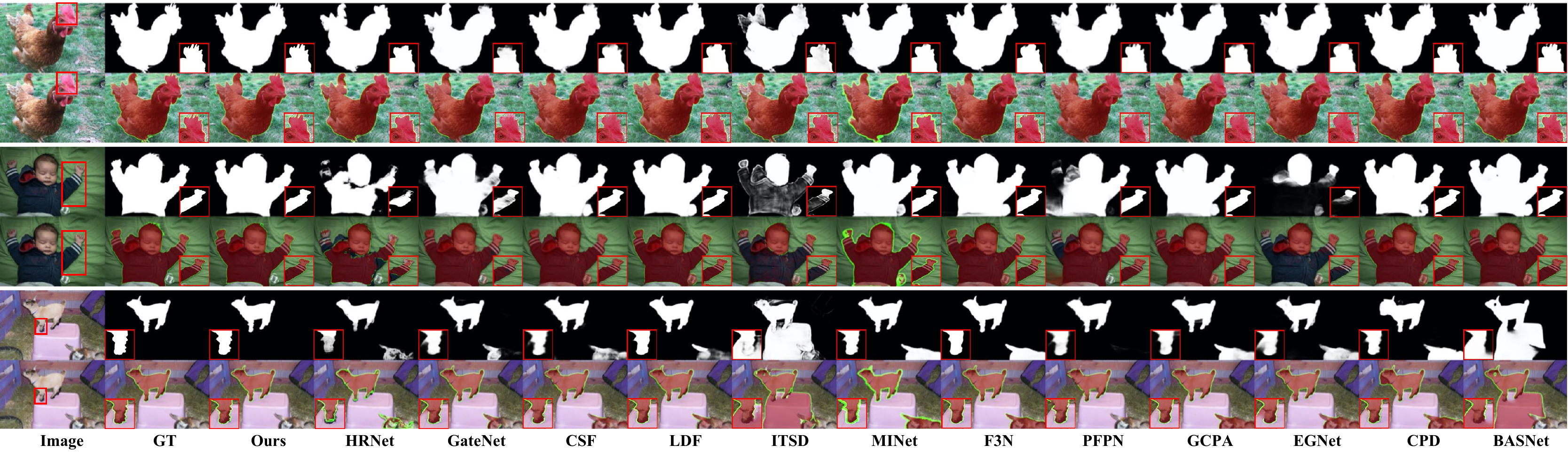}
\caption{Visual comparison between our method and other SOTA methods. Each sample occupies two rows. Best viewed by zooming in. It can be clearly observed that our method achieves impressive performance in all these cases.}
\label{Visresults}
\vspace{-0.5cm}
\end{figure*}

\section{Experiment}
\subsection{Experimental Settings} 
\textbf{Implementation Details}. Following the works~\cite{DBLP:conf/cvpr/QinZHGDJ19,DBLP:conf/iccv/ZhaoLFCYC19,DBLP:conf/cvpr/PangZZL20,DBLP:conf/cvpr/WeiWWSH020}, we train our proposed network on DUTS-TR. We use Pytorch\footnote{https://pytorch.org/} to implement our model.
A GTX 1080Ti GPU is used for acceleration. VGG-16~\cite{DBLP:journals/corr/SimonyanZ14a} is used as the backbone network of LRSCN, and the whole network is trained end-to-end by stochastic gradient descent (SGD). For a more comprehensive demonstration, we also trained our network with ResNet-50~\cite{DBLP:conf/cvpr/HeZRS16} backbone. Maximum learning rate is set to 0.001 for backbone and 0.01 for other parts. Warm-up and linear decay strategies are used to adjust the learning rate. Momentum and weight decay are set to 0.9 and 0.0005 respectively. Batchsize is set to 32 and maximum epoch is set to 100. Horizontal flip and multi-scale input images are utilized for data augmentation as done in ~\cite{DBLP:conf/cvpr/QinZHGDJ19,DBLP:conf/iccv/ZhaoLFCYC19,DBLP:journals/corr/abs-1911-11445}. During testing, the input of LRSCN is about $352 \times 352$ resolution.

The learning rate of HRRN is initialized to 0.0005. Warmup and cosine decay are applied to the learning rate. The network HRRN is trained for 10000 iterations with a batch size of 20. During training, the resolution of input image and trimap is $512 \times 512$. During testing, we first resize the image and trimap to $1024 \times 1024$, then we split the image and trimap into four sub-images and sub-trimaps with $512 \times 512$ resolution, as shown in Fig.\ref{framework}. Finally, we send each sub-image and sub-trimap together to HRRN to generate sub-prediction result, and use 4 sub-predictions stitched together to make one high-resolution saliency result.

\noindent\textbf{Evaluation Datasets}. 
Following work~\cite{DBLP:conf/iccv/ZengZLZL19}, we evaluate our method on two high-resolution saliency detection datasets, including HRSOD-TE and DAVIS-S, which contain 400 and 92 images. DAVIS-S dataset is collected from DAVIS~\cite{DBLP:conf/cvpr/PerazziPMGGS16}. Images in these two datasets are precisely annotated and have very high resolutions (i.e.,$1920 \times 1080$). We also evaluate our method on three low-resolution datasets, including DUT-OMRON~\cite{DBLP:conf/cvpr/YangZLRY13}, DUTS-TE~\cite{DBLP:conf/cvpr/WangLWF0YR17} and HKU-IS~\cite{DBLP:conf/cvpr/LiY15}, which contain 5168, 5019 and 4447 images. Our results are available at \url{https://github.com/luckybird1994/HQSOD}.

\noindent\textbf{Evaluation Metrics}. Six metrics are used to evaluate the performance of our method. The first is Mean Absolute Error (MAE), which characterize the average $1$-norm distance between ground truth maps and predictions. The second is F-measure ($F_{\beta}$ and $F_{\beta}^{max}$), a weighted mean of average precision and average recall, calculated by
$F_{\beta}=\frac{(1+\beta^{2}) \times Precision \times Recall}{\beta^{2} \times Precision + Recall}$. We set $\beta^{2}$ to be 0.3 as suggested in \cite{DBLP:journals/tip/BorjiCJL15}. The third is Structure Measure ($S_m$), a metric to evaluate the spatial structure similarities of saliency maps based on both region-aware structural similarity $S_{r}$ and object-aware structural similarity $S_{o}$, defined as $S_{\alpha} = \alpha \ast S_{r} + (1- \alpha) \ast S_{o}$, where $\alpha = 0.5$~\cite{DBLP:conf/iccv/FanCLLB17}. In addition, precision-recall (PR) curve is used to show the whole performance. To further evaluate the boundary quality, Following ~\cite{DBLP:conf/iccv/ZengZLZL19} and ~\cite{DBLP:conf/cvpr/ZhangYLSLD20}, we use Boundary Displacement Error (BDE)~\cite{DBLP:conf/eccv/FreixenetMRMC02} and $B_\mu$ metrics. More details about BDE and $B_\mu$ can be found in section 7 of supplementary materials. The last two metrics are only used in two high-resolution datasets, because their boundaries annotation is accurate and evaluating results are reliable.

\subsection{Comparisons with the State-of-the-Arts}
We compare our approach with 16
SOTA methods, including Amulet~\cite{DBLP:conf/iccv/ZhangWLWR17}, R3Net~\cite{DBLP:conf/ijcai/DengHZXQHH18}, DGRL~\cite{DBLP:conf/cvpr/WangZWL0RB18}, 
DSS~\cite{DBLP:journals/pami/HouCHBTT19}, 
BASNet~\cite{DBLP:conf/cvpr/QinZHGDJ19},
CPD~\cite{DBLP:conf/cvpr/WuSH19},  EGNet~\cite{DBLP:conf/iccv/ZhaoLFCYC19}, PFPN~\cite{DBLP:conf/aaai/WangCZZ0G20}, GCPA~\cite{DBLP:conf/aaai/ChenXCH20}, F3N~\cite{DBLP:journals/corr/abs-1911-11445}, MINet~\cite{DBLP:conf/cvpr/PangZZL20}, ITSD~\cite{DBLP:conf/cvpr/ZhouXLCY20}, LDF~\cite{DBLP:conf/cvpr/WeiWWSH020}, GateNet~\cite{DBLP:journals/corr/abs-2007-08074}, CSF~\cite{DBLP:journals/corr/abs-2003-05643} and HRNet~\cite{DBLP:conf/iccv/ZengZLZL19}. For a fair comparison, we use either the implementations with recommended parameter settings or the saliency maps provided by the authors. The evaluation toolbox used in this paper is same as F3N~\cite{DBLP:journals/corr/abs-1911-11445}. 

\textbf{Quantitative Evaluation.}
From Table.\ref{total}, when we train our network only using DUTS (Ours), our method can already improve the $F_\beta^{max}$,$F_\beta$, $S_m$ and MAE achieved by the best-performing existing algorithms, especially two high-resolution test datasets. It is worth noting that for boundary accuracy, 
our method is far better than other methods on two high-resolution. These results demonstrate the efficiency of the proposed disentangled SOD framework in both identifying the salient regions and estimating the accurate objects boundaries. Other than numerical results, we also show the PR curves on two high-resolution datasets and three low-resolution datasets in Fig.\ref{PR}. As can be seen, the PR curves by our method (red ones) are especially outstanding compared to all other previous methods. Besides, shorter PR curves imply that our saliency maps are usually more assertive with sharper boundaries than the results of other methods.
An interesting observation is that when we add HRSOD-training datasets (resized to low-resolution like $352 \times 352$) in LRSCN, the performance in two high-resolution datasets HRSOD-TE and DAVIS-S can be further improved. However, this practice seems to be of little help in improving the performance in other three low-resolution datasets. A similar phenomenon can also be found in the performance of HRNet~\cite{DBLP:conf/iccv/ZengZLZL19}. 
We think there could be some image selection or data annotation biases between the high-resolution datasets and the low-resolution datasets, which cause this phenomenon.

\textbf{Qualitative Evaluation.}
To exhibit the superiority of the proposed approach, Fig.\ref{Visresults} shows representative examples of saliency maps generated by our approach and other state-of-the-art algorithms. As can be seen, with the help of LRSCN, our method can not only keep the wholeness of the salient object (row 3), but also accurately locate salient objects and suppress non-salient regions (row 5), compared to other methods. HRRN can help the model to restore accurate and complete boundaries of salient objects, which are 
more consistent with the GT boundaries. It can be clearly observed that our method achieves impressive performance in all these cases, which indicates the eﬀectiveness of disentangled framework and uncertainty loss. More comparison experiments can be found in section 2 of supplementary materials.

\begin{table}
\centering
\caption{Ablation Studies of LRSCN.}
\scalebox{0.4}{
\begin{tabular}{l|cccccc|cccccc}
\hline
\multicolumn{1}{c|}{}                                &                              & \multicolumn{5}{c|}{HRSOD-TE}                                                                                                                             &                              & \multicolumn{5}{c}{DAVIS-S}                                                                                                                               \\ \cline{2-13} 
\multicolumn{1}{c|}{\multirow{-2}{*}{Conﬁgurations}} & $F_\beta^{max}$              & $F_\beta$                    & $S_m$                        & MAE                          & BDE                           & $B_\mu$                      & $F_\beta^{max}$              & $F_\beta$                    & $S_m$                        & MAE                          & BDE                           & $B_\mu$                      \\ \hline
Baseline+HRRN                                        & 0.900                        & 0.880                        & 0.896                        & 0.034                        & 65.732                        & 0.842                        & 0.899                        & 0.887                        & 0.919                        & 0.021                        & 31.201                        & 0.678                        \\
Baseline+ME+HRRN                                     & 0.910                        & 0.894                        & 0.902                        & 0.031                        & 60.040                        & 0.801                        & 0.920                        & 0.904                        & 0.924                        & 0.019                        & 24.022                        & 0.612                        \\
Baseline+CF+HRRN                                     & 0.909                        & 0.892                        & 0.900                        & 0.030                        & 59.028                        & 0.804                        & 0.918                        & 0.905                        & 0.923                        & 0.018                        & 23.988                        & 0.605                        \\
Baseline+MECF+HRRN                                   & 0.913                        & 0.898                        & 0.909                        & 0.029                        & 54.377                        & 0.766                        & 0.928                        & 0.915                        & 0.929                        & 0.016                        & 20.010                        & 0.578                        \\
Baseline+MECF+SGA+HRRN                               & {\color[HTML]{FE0000} 0.918} & {\color[HTML]{FE0000} 0.902} & {\color[HTML]{FE0000} 0.912} & {\color[HTML]{FE0000} 0.027} & {\color[HTML]{FE0000} 48.468} & {\color[HTML]{FE0000} 0.711} & {\color[HTML]{FE0000} 0.933} & {\color[HTML]{FE0000} 0.919} & {\color[HTML]{FE0000} 0.933} & {\color[HTML]{FE0000} 0.015} & {\color[HTML]{FE0000} 15.676} & {\color[HTML]{FE0000} 0.536} \\ \hline
\end{tabular}}
\label{LRSCN}
\end{table}

\begin{table}
\centering
\caption{Ablation Studies of HRRN.}
\scalebox{0.38}{
\begin{tabular}{c|l|cccccc|cccccc}
\hline
                                & \multicolumn{1}{c|}{}                                 & \multicolumn{6}{c|}{HRSOD-TE}                                                                                                                                                            & \multicolumn{6}{c}{DAVIS-S}                                                                                                                                                              \\ \cline{3-14} 
\multirow{-2}{*}{Ablation}      & \multicolumn{1}{c|}{\multirow{-2}{*}{Configurations}} & $F_{\beta}^{max}$              & $F_{\beta}$                    & $S_{m}$                        & MAE                          & BDE                           & $B_{\mu}$                      & $F_{\beta}^{max}$              & $F_\beta$                    & $S_m$                        & MAE                          & BDE                           & $B_\mu$                      \\ \hline
                                & Ours($L1+L_{uncertainy}$)                             & 0.918                        & 0.902                        & 0.912                        & 0.027                        & 48.468                        & 0.711                        & 0.933                        & 0.919                        & 0.933                        & 0.015                        & 15.676                        & 0.536                        \\
                                & Ours($L1$)                                            & 0.907                        & 0.896                        & 0.908                        & 0.029                        & 53.891                        & 0.780                        & 0.921                        & 0.909                        & 0.927                        & 0.016                        & 18.014                        & 0.622                        \\
\multirow{-3}{*}{\rotatebox{90}{Loss}}          & Ours-DH($L1+L_{uncertainy}$)                        & 0.918                        & 0.901                        & 0.911                        & 0.027                        & 48.878                        & 0.712                        & 0.933                        & 0.920                        & 0.934                        & 0.015                        & 17.670                        & 0.540                        \\ \hline
                                & Ours(LRSCN)                                           & 0.898                        & 0.885                        & 0.899                        & 0.034                        & 64.805                        & 0.822                        & 0.909                        & 0.898                        & 0.920                        & 0.022                        & 28.798                        & 0.684                        \\
                                & Ours(LRSCN+HRRN)                                      & {\color[HTML]{FE0000} 0.918} & {\color[HTML]{FE0000} 0.902} & {\color[HTML]{FE0000} 0.912} & {\color[HTML]{FE0000} 0.027} & {\color[HTML]{FE0000} 48.468} & {\color[HTML]{FE0000} 0.711} & {\color[HTML]{FE0000} 0.933} & {\color[HTML]{FE0000} 0.919} & {\color[HTML]{FE0000} 0.933} & {\color[HTML]{FE0000} 0.015} & {\color[HTML]{FE0000} 15.676} & {\color[HTML]{FE0000} 0.536} \\
                                & Ours(LRSCN+CRF)                                      & 0.905                        & 0.896                        & 0.897                        & 0.029                        & 60.521                        & 0.797                        & 0.920                        & 0.907                        & 0.918                        & 0.018                        & 24.455                        & 0.665                        \\ \cline{2-14} 
                                & EGNet                                                 & 0.883                        & 0.814                        & 0.888                        & 0.044                        & 73.500                        & 0.896                        & 0.886                        & 0.794                        & 0.897                        & 0.030                        & 37.369                        & 0.799                        \\
                                & EGNet(+HRRN)                                         & {\color[HTML]{FE0000} 0.900}   & {\color[HTML]{FE0000} 0.862} & {\color[HTML]{FE0000} 0.889} & {\color[HTML]{FE0000} 0.039} & {\color[HTML]{FE0000} 72.982} & {\color[HTML]{FE0000} 0.753} & {\color[HTML]{FE0000} 0.904} & {\color[HTML]{FE0000} 0.858} & {\color[HTML]{FE0000} 0.898} & {\color[HTML]{FE0000} 0.024} & {\color[HTML]{FE0000} 34.860} & {\color[HTML]{FE0000} 0.603} \\
                                & EGNet(+CRF)                                           & 0.895                        & 0.858                        & 0.882                        & 0.039                        & 73.348                        & 0.796                        & 0.902                        & 0.846                        & 0.892                        & 0.025                        & 34.721                        & 0.679                        \\ \cline{2-14} 
                                & CPD                                                   & 0.876                        & 0.829                        & 0.887                        & 0.039                        & 72.686                        & 0.824                        & 0.878                        & 0.822                        & 0.903                        & 0.025                        & 36.649                        & 0.703                        \\
                                & CPD(+HRRN)                                            & {\color[HTML]{FE0000} 0.891} & {\color[HTML]{FE0000} 0.855} & {\color[HTML]{FE0000} 0.888} & {\color[HTML]{FE0000} 0.036} & {\color[HTML]{FE0000} 72.796} & {\color[HTML]{FE0000} 0.734} & {\color[HTML]{FE0000} 0.893} & {\color[HTML]{FE0000} 0.859} & {\color[HTML]{FE0000} 0.907} & {\color[HTML]{FE0000} 0.022} & {\color[HTML]{FE0000} 35.294} & {\color[HTML]{FE0000} 0.564} \\
                                & CPD(+CRF)                                             & 0.885                        & 0.851                        & 0.884                        & 0.037                        & 76.440                        & 0.772                        & 0.884                        & 0.852                        & 0.903                        & 0.023                        & 36.915                        & 0.633                        \\ \cline{2-14} 
                                & BASNet(En-De)                                         & 0.873                        & 0.827                        & 0.888                        & 0.039                        & 70.944                        & 0.824                        & 0.852                        & 0.802                        & 0.880                        & 0.039                        & 48.309                        & 0.703                        \\
                                & BASNet(En-De+HRRN)                                    & {\color[HTML]{FE0000} 0.895} & {\color[HTML]{FE0000} 0.858} & {\color[HTML]{FE0000} 0.892} & {\color[HTML]{FE0000} 0.036} & {\color[HTML]{FE0000} 67.191} & {\color[HTML]{FE0000} 0.719} & {\color[HTML]{FE0000} 0.875} & {\color[HTML]{FE0000} 0.832} & {\color[HTML]{FE0000} 0.882} & {\color[HTML]{FE0000} 0.036} & {\color[HTML]{FE0000} 45.922} & {\color[HTML]{FE0000} 0.566} \\
                                & BASNet(En-De+RRM)                                     & 0.878                        & 0.831                        & 0.890                        & 0.038                        & 67.643                        & 0.823                        & 0.857                        & 0.806                        & 0.881                        & 0.039                        & 46.283                        & 0.705                        \\
\multirow{-13}{*}{\rotatebox{90}{Architecture}} & BASNet(En-De+CRF)                                     & 0.888                        & 0.850                        & 0.886                        & 0.036                        & 77.235                        & 0.758                        & 0.867                        & 0.825                        & 0.878                        & 0.037                        & 46.423                        & 0.629                        \\ \hline
\end{tabular}}
\label{HRRN}
\end{table}

\begin{figure}
\centering
\includegraphics[scale=0.25]{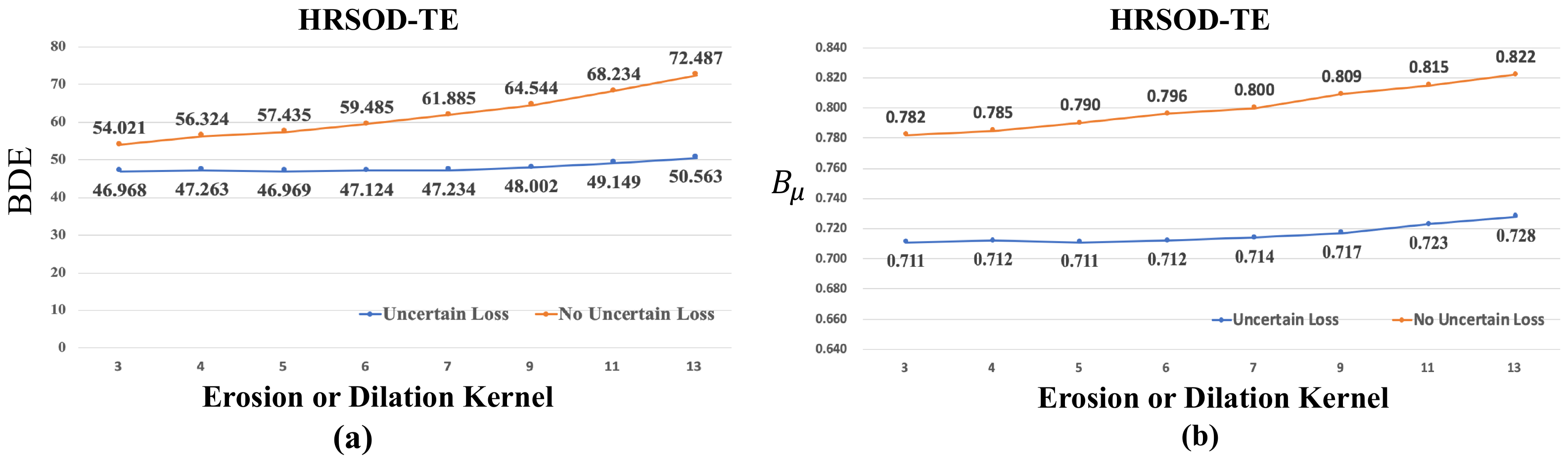}
\caption{Training the network with noisy data.}
\label{Fig8}
\vspace{-0.5cm}
\end{figure}

\subsection{Ablation Studies}
To validate the effectiveness of the proposed components of our method, we conduct a series of experiments on two high-resolution datasets with different settings under VGG-16 backbone. Specifically, we first verify the effectiveness of MECF and SGA in LRSCN. Then we validate the  effectiveness of uncertainty loss in HRRN and the superiority of the proposed disentangled architecture.

\textbf{Ablation Studies of LRSCN.} To prove the effectiveness of MECF and SGA module, we report the quantitative comparison results of LRSCN with different architectures in Table.\ref{LRSCN}. Baseline denotes that we conduct a experiment over on $L_{LRSCN}$ with a pure U-Net architecture. We can see that only using ME or CF can already heavily improve the performance. A better performance has been achieved through the combination of these two architectures. Finally, performace can be further improved by SGA module, especially BDE and $B_\mu$, which means that SGA can help 
generate accurate trimap. While $L_{saliency}$ is not our core innovation, more ablation studies about $L_{saliency}$ can be found in section 5 of supplementary materials.

\textbf{Ablation Studies of HRRN.} In HRRN, uncertainty loss play a key role to estimate the accurate objects boundaries, so we first investigate the eﬀectiveness of our proposed uncertainty loss. From Table.\ref{HRRN}, we can see that without uncertainty loss (Ours($L_1$)), the performance decreased a lot. Besides, when we add high-resolution HRSOD-Training datasets in HRRN (Ours-DH($L_1+L_{uncertainty}$)), the performance has no obvious improvement, which demonstrates that 
our network is not reliant on accurately annotated high-resolution images during training. To further demonstrate the effectiveness of our disentangled framework, we compare our HRRN with CRF~\cite{DBLP:conf/nips/KrahenbuhlK11}, a widely used post-processing for saliency detection. Results in Table.3 show that our proposed method (Ours (LRSCN + HRRN)) outperforms CRF (Ours (LRSCN + CRF)) by a large margin. The same phenomenon can be found in the refinement of EGNet, CPD and BASNet (Their trimaps are generated with the corresponding saliency maps following Eq.1). Moreover, compared to RRM module proposed in BASNet, our HRRN can better improve performance. This ablation study demonstrates the superiority of HRRN within our novel disentangled framework. More analyses of the proposed disentangled framework can be found in section 3 of supplementary materials.

\textbf{Analysis of Uncertainty Loss}
To further demonstrate that uncertainty loss can make network more robust to noisy data, we erase and dilate binary groundtruth maps of DUT-TR at the object boundaries with a random pixel number (3,4,5,6,7,9,11,13) to generate noisy training data. Then we train the network on these noisy data, BDE and $B_\mu$ results on HRSOD-TE are reported in Fig.\ref{Fig8}(a) and Fig.\ref{Fig8}(b). When the erosion or dilation kernel ranges from 3 to 7, the network trained with uncertainty loss has a rather stable  performance. With the increase in erosion or dilation kernels, even though the performance is dropped, training with uncertainty loss still yields a better model than training without uncertainty loss. This experiment further validates the effectiveness of uncertainty loss.

\section{Conclusions}
In this paper, we argue that there are two difficult and inherently different problems in high-resolution SOD. From this perspective, we propose a novel deep learning framework to disentangle the high-resolution SOD into two tasks: LRSCN and HRRN. LRSCN can identify the definite salient, background and uncertain  regions at low-resolution with sufficient semantics. While HRRN can accurately refining the saliency value of pixels in the uncertain region to preserve a clear object boundary at high-resolution with limited GPU memory. We also make the earliest efforts to introduce the uncertainty into SOD network training, which empower HRRN to learn rich details without using any high-resolution training datasets. Extensive evaluations on high-resolution datasets and popular benchmark datasets not only verify the  superiority of our method but also demonstrate the importance of disentanglement for SOD. We 
believe our novel disentanglement view in this work can contribute to other high-resolution computer vision tasks in the future.

\clearpage

\begin{appendices}

\section{More Quantitative and Qualitative Results}
\subsection{Quantitative Comparison on more datasets}

\begin{figure*}[!htp]
\centering
\includegraphics[scale=0.73]{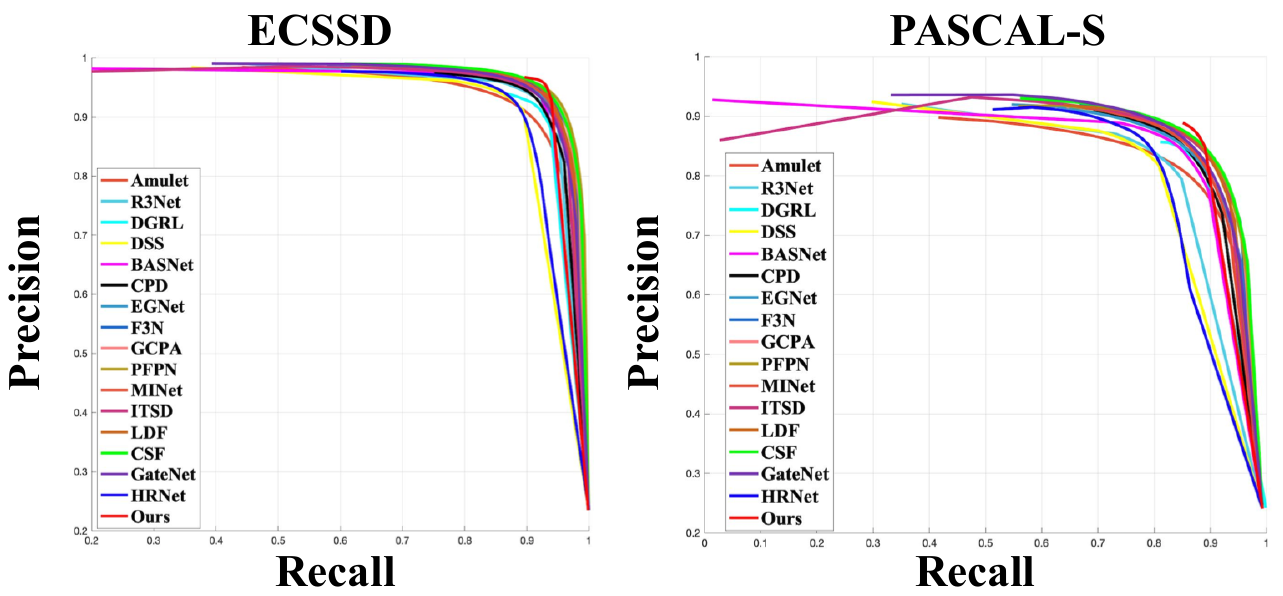}
\caption{Comparison of PR curves across another two conventional low-resolution datasets.}
\label{ecssd_pr}
\end{figure*}

We compare our method with other SOTA methods on another two conventional low-resolution datasets ECSSD~\cite{DBLP:journals/pami/ShiYXJ16} and PASCAL-S~\cite{DBLP:conf/cvpr/LiHKRY14}, which have 1000 and 850 images respectively. The results are reported in Table.\ref{ecssd}. It can be seen that our method consistently outperforms other methods across these two conventional datasets. We also show their PR curves in Fig.\ref{ecssd_pr}. It should be noted that $F_{max}$ represents $F^{max}_{\beta}$. We apologize for this writing error of Table.2 in the main text.

\begin{figure*}[!t]
\centering
\includegraphics[scale=0.6]{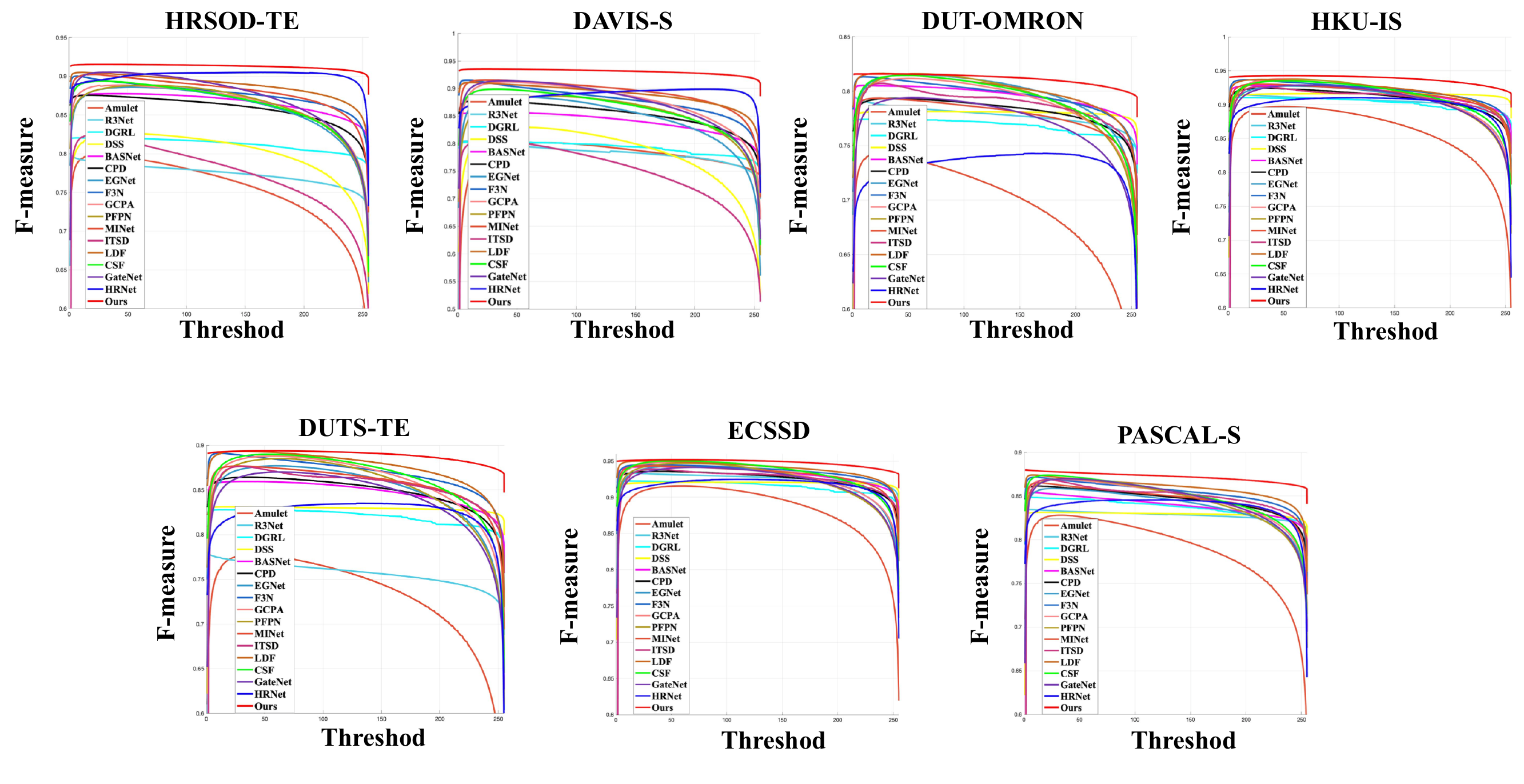}
\caption{Comparison of the F-measure curves across on two high-resolution and five low-resolution datasets.}
\label{Fig3}
\end{figure*}

F-measure curves of different methods are displayed in Fig.\ref{Fig3}, for overall comparisons. One can observe that our approach noticeably outperforms all the other state-of-the-art methods. These observations demonstrate the efficiency and robustness of our proposed method across various challenging datasets.

SOC~\cite{DBLP:conf/eccv/FanCLGHB18} is a new challenging dataset with nine attributes. In Table.\ref{soc}, we evaluate the mean F-measure score of our method as well as 11 state-of-the-art methods. We can see the proposed model achieves the competitive results among most of attributes and the overall score is best.

Model size and running time comparisons among different methods are also reported in Table.\ref{modelsize}. It can be seen that with the high-resolution input, our method is more efficient than HRNet. For fair, the running time analysis of our method is also conducted with the low-resolution input ($352\times352$), and our method runs at a competitive efficiency.

\begin{table*}
\centering
\caption{Quantitative comparison with SOTA methods on another two conventional datasets.}
\scalebox{0.7}{
\begin{tabular}{llllllllll}
\hline
\multicolumn{1}{c|}{}                         & \multicolumn{1}{c|}{}                                                                              & \multicolumn{4}{c|}{ECSSD}                                                                                                                     & \multicolumn{4}{c}{PASCAL-S}                                                                                                \\ \cline{3-10} 
\multicolumn{1}{c|}{\multirow{-2}{*}{Models}} & \multicolumn{1}{c|}{\multirow{-2}{*}{\begin{tabular}[c]{@{}c@{}}Training\\ datasets\end{tabular}}} & $F_\beta^{max}$              & $F_\beta$                    & $S_m$                        & \multicolumn{1}{c|}{MAE}                          & $F_\beta^{max}$              & $F_\beta$                    & $S_m$                        & MAE                          \\ \hline
\multicolumn{10}{c}{VGG-16 backbone}                                                                                                                                                                                                                                                                                                                                                                                            \\ \hline
\multicolumn{1}{c|}{Amulet(ICCV2017)}         & \multicolumn{1}{c|}{MK}                                                                            & 0.915                        & 0.868                        & 0.894                        & \multicolumn{1}{c|}{0.059}                        & 0.828                        & 0.757                        & 0.818                        & 0.100                        \\
\multicolumn{1}{c|}{DGRL(CVPR2018)}           & \multicolumn{1}{c|}{DUTS}                                                                          & 0.922                        & 0.903                        & 0.906                        & \multicolumn{1}{c|}{0.043}                        & 0.849                        & 0.807                        & 0.834                        & 0.074                        \\
\multicolumn{1}{c|}{DSS(TPAMI2019)}           & \multicolumn{1}{c|}{MB}                                                                            & 0.921                        & 0.904                        & 0.882                        & \multicolumn{1}{c|}{0.052}                        & 0.831                        & 0.802                        & 0.798                        & 0.094                        \\
\multicolumn{1}{c|}{CPD(CVPR2019)}            & \multicolumn{1}{c|}{DUTS}                                                                          & 0.936                        & 0.917                        & {\color[HTML]{32CB00} 0.917} & \multicolumn{1}{c|}{{\color[HTML]{3166FF} 0.037}} & 0.861                        & {\color[HTML]{3166FF} 0.824}                        & 0.842                        & 0.072                        \\
\multicolumn{1}{c|}{EGNET(ICCV2019)}          & \multicolumn{1}{c|}{DUTS}                                                                          & {\color[HTML]{32CB00}0.943} & 0.913                        & 0.913                        & \multicolumn{1}{c|}{0.041}                        & 0.858                        & 0.809                        & 0.848                        & 0.077                        \\
\multicolumn{1}{c|}{MINet(CVPR2020)}          & \multicolumn{1}{c|}{DUTS}                                                                          & 
{\color[HTML]{32CB00} 0.943} & {\color[HTML]{32CB00} 0.922} & {\color[HTML]{32CB00} 0.917} & \multicolumn{1}{c|}{{\color[HTML]{32CB00} 0.036}} & 0.865                        & {\color[HTML]{32CB00} 0.829} & {\color[HTML]{FE0000} 0.854} & {\color[HTML]{3166FF} 0.064} \\
\multicolumn{1}{c|}{ITSD(CVPR2020)}           & \multicolumn{1}{c|}{DUTS}                                                                          & 0.939                        & 0.875                        & {\color[HTML]{3166FF} 0.914} & \multicolumn{1}{c|}{0.040}                        & 0.869                        & 0.773                        & {\color[HTML]{32CB00} 0.853}                        & {0.068} \\
\multicolumn{1}{c|}{GateNet(ECCV2020)}        & \multicolumn{1}{c|}{DUTS}                                                                          & {\color[HTML]{3166FF} 0.941}                        & 0.896                        & {\color[HTML]{32CB00} 0.917} & \multicolumn{1}{c|}{0.041}                        & {\color[HTML]{3166FF} 0.870} & 0.797                        & {\color[HTML]{32CB00} 0.853}                        & {0.068} \\
\multicolumn{1}{c|}{HRNet(ICCV2019)}          & \multicolumn{1}{c|}{DUTS+HR}                                                                       & 0.925                        & 0.905                        & 0.888                        & \multicolumn{1}{c|}{0.052}                        & 0.846                        & 0.804                        & 0.817                        & 0.079                        \\
\multicolumn{1}{c|}{Ours}                     & \multicolumn{1}{c|}{DUTS}                                                                          & {\color[HTML]{FE0000} 0.948} & {\color[HTML]{FE0000} 0.931} & {\color[HTML]{FE0000} 0.918} & \multicolumn{1}{c|}{{\color[HTML]{FE0000} 0.034}} & {\color[HTML]{FE0000} 0.874} & {\color[HTML]{FE0000} 0.845} & {\color[HTML]{FE0000} 0.854} & {\color[HTML]{32CB00} 0.063} \\
\multicolumn{1}{c|}{Ours-DH}                  & \multicolumn{1}{c|}{DUTS+HR-L}                                                                     & {0.938} & 
{\color[HTML]{3166FF} 0.918} & {0.904} & \multicolumn{1}{c|}{{0.040}} & {\color[HTML]{32CB00} 0.871} & {\color[HTML]{FE0000} 0.845} & {\color[HTML]{3166FF} 0.851} & {\color[HTML]{FE0000} 0.061} \\ \hline
\multicolumn{10}{c}{ResNet-50/ResNet-101/ResNeXt-101/Res2Net50 backbone}                                                                                                                                                                                                                                                                                                                                                      \\ \hline
\multicolumn{1}{c|}{R3Net(IJCAI2018)}         & \multicolumn{1}{c|}{MK}                                                                            & 0.934                        & 0.883                        & 0.910                        & \multicolumn{1}{c|}{0.051}                        & 0.834                        & 0.775                        & 0.809                        & 0.101                        \\
\multicolumn{1}{c|}{BasNet(CVPR2019)}         & \multicolumn{1}{c|}{DUTS}                                                                          & 0.942                        & 0.880                        & 0.916                        & \multicolumn{1}{c|}{0.037}                        & 0.854                        & 0.775                        & 0.832                        & 0.076                        \\
\multicolumn{1}{c|}{PFPN(AAAI2020)}           & \multicolumn{1}{c|}{DUTS}                                                                          & 0.947                        & 0.917                        & {\color[HTML]{32CB00} 0.927} & \multicolumn{1}{c|}{0.035}                        & 0.870                        & 0.824                        & 0.851                        & 0.065                        \\
\multicolumn{1}{c|}{GCPA(AAAI2020)}           & \multicolumn{1}{c|}{DUTS}                                                                          & 0.948                        & 0.919                        & {\color[HTML]{32CB00} 0.927} & \multicolumn{1}{c|}{0.035}                        & 0.869                        & 0.827                        & {\color[HTML]{32CB00} 0.860} & 0.062                        \\
\multicolumn{1}{c|}{F3N(AAAI2020)}            & \multicolumn{1}{c|}{DUTS}                                                                          & 0.945                        & {\color[HTML]{3166FF} 0.925} & 0.924                        & \multicolumn{1}{c|}{0.036}                        & 0.872                        & {\color[HTML]{3166FF} 0.840} & 0.855                        & 0.062                        \\
\multicolumn{1}{c|}{LDF(CVPR2020)}            & \multicolumn{1}{c|}{DUTS}                                                                          & 0.950                        & {\color[HTML]{32CB00} 0.930} & 0.924                        & \multicolumn{1}{c|}{0.034}                        & {\color[HTML]{3166FF} 0.874} & {\color[HTML]{32CB00} 0.843} & {\color[HTML]{3166FF} 0.859} & {\color[HTML]{3166FF} 0.061} \\
\multicolumn{1}{c|}{CSF(ECCV2020)}            & \multicolumn{1}{c|}{DUTS}                                                                          & {\color[HTML]{3166FF} 0.950} & {\color[HTML]{3166FF} 0.925} & {\color[HTML]{32CB00} 0.927} & \multicolumn{1}{c|}{{\color[HTML]{3166FF} 0.033}} & {\color[HTML]{3166FF} 0.874} & 0.823                        & 0.858                        & 0.069                        \\
\multicolumn{1}{c|}{Ours}                     & \multicolumn{1}{c|}{DUTS}                                                                          & {\color[HTML]{32CB00} 0.952} & {\color[HTML]{FE0000} 0.941} & {\color[HTML]{FE0000} 0.928} & \multicolumn{1}{c|}{{\color[HTML]{FE0000} 0.029}} & {\color[HTML]{FE0000} 0.880} & {\color[HTML]{FE0000} 0.852} & {\color[HTML]{FE0000} 0.861} & {\color[HTML]{FE0000} 0.059} \\
\multicolumn{1}{c|}{Ours-DH}                  & \multicolumn{1}{c|}{DUTS+HR-L}                                                                     & {\color[HTML]{FE0000} 0.953} & {\color[HTML]{FE0000} 0.941} & {\color[HTML]{3166FF} 0.926} & \multicolumn{1}{c|}{{\color[HTML]{32CB00} 0.030}} & {\color[HTML]{32CB00} 0.878} & {\color[HTML]{FE0000} 0.852} & {\color[HTML]{3166FF} 0.859} & {\color[HTML]{32CB00} 0.060} \\ \hline
\end{tabular}}
\label{ecssd}
\end{table*}

\begin{table*}
\centering
\caption{Performance on SOC of different attributes. The last row shows the whole performance on the SOC dataset.}
\scalebox{0.63}{
\begin{tabular}{cccccclccccccc}
\hline
Attr & BASNet & CPD                          & EGNet & F3N                          & GCPA                         & PFPN                         & ITSD  & LDF                          & MINet                        & CSF   & GateNet & Ours                         & \multicolumn{1}{l}{Ours-DH}  \\ \hline
AC   & 0.723  & 0.750                        & 0.756 & 0.784                        & 0.780                        & 0.772                        & 0.611 & {\color[HTML]{FE0000} 0.796} & {\color[HTML]{3166FF} 0.790} & 0.730 & 0.748   & {\color[HTML]{32CB00} 0.793} & 0.788                        \\
BO   & 0.511  & 0.794                        & 0.702 & 0.791                        & {\color[HTML]{FE0000} 0.882} & {\color[HTML]{333333} 0.837} & 0.499 & 0.807                        & 0.814                        & 0.825 & 0.737   & {\color[HTML]{32CB00} 0.858} & {\color[HTML]{3166FF} 0.848} \\
CL   & 0.682  & {\color[HTML]{32CB00} 0.771} & 0.726 & 0.757                        & 0.765                        & 0.765                        & 0.610 & 0.763                        & {\color[HTML]{3166FF} 0.770} & 0.751 & 0.754   & {\color[HTML]{FE0000} 0.789} & {\color[HTML]{FE0000} 0.789} \\
HO   & 0.772  & 0.777                        & 0.756 & 0.790                        & 0.780                        & 0.777                        & 0.685 & {\color[HTML]{32CB00} 0.797} & {\color[HTML]{3166FF} 0.792} & 0.779 & 0.788   & {\color[HTML]{FE0000} 0.817} & {\color[HTML]{FE0000} 0.817} \\
MB   & 0.687  & 0.715                        & 0.687 & {\color[HTML]{3166FF} 0.761} & 0.691                        & 0.705                        & 0.589 & {\color[HTML]{333333} 0.758} & 0.708                        & 0.702 & 0.725   & {\color[HTML]{32CB00} 0.764} & {\color[HTML]{FE0000} 0.768} \\
OC   & 0.686  & 0.719                        & 0.702 & 0.724                        & 0.720                        & 0.729                        & 0.629 & {\color[HTML]{32CB00} 0.739} & {\color[HTML]{3166FF} 0.729} & 0.703 & 0.728   & {\color[HTML]{FE0000} 0.771} & {\color[HTML]{FE0000} 0.771} \\
OV   & 0.720  & 0.764                        & 0.764 & 0.793                        & {\color[HTML]{32CB00} 0.802} & 0.806                        & 0.639 & {\color[HTML]{FE0000} 0.805} & 0.788                        & 0.772 & 0.787   & {\color[HTML]{3166FF} 0.798} & {\color[HTML]{32CB00} 0.802} \\
SC   & 0.708  & 0.723                        & 0.683 & {\color[HTML]{3166FF} 0.747} & 0.707                        & 0.697                        & 0.592 & {\color[HTML]{333333} 0.746} & 0.726                        & 0.690 & 0.715   & {\color[HTML]{FE0000} 0.785} & {\color[HTML]{32CB00} 0.782} \\
SO   & 0.632  & 0.643                        & 0.614 & {\color[HTML]{3166FF} 0.668} & 0.640                        & 0.636                        & 0.523 & {\color[HTML]{32CB00} 0.691} & 0.652                        & 0.621 & 0.641   & {\color[HTML]{FE0000} 0.713} & {\color[HTML]{FE0000} 0.713} \\ \hline
Avg  & 0.680  & 0.740                        & 0.710 & {\color[HTML]{333333} 0.757} & 0.752                        & 0.747                        & 0.597 & {\color[HTML]{3166FF} 0.767} & 0.753                        & 0.730 & 0.736   & {\color[HTML]{FE0000} 0.788}    & {\color[HTML]{32CB00} 0.787} \\ \hline
\end{tabular}}
\label{soc}
\end{table*}

\begin{table*}
\centering
\caption{Model size and running time comparisons between our approach and SOTA methods.}
\scalebox{0.57}{
\begin{tabular}{c|ccccccccc}
\hline
                      & Ours             & Ours           & DGRL           & DSS            & BASNet         & EGNet          & GCPA           & PFPN           & R3Net          \\ \hline
Model Size(MB)        & 309.6            & 309.6          & 648            & 447.3          & 412.2          & 332.1          & 255.8          & 243.0          & 214.2          \\ \hline
Time(s)               & 0.21             & 0.05           & 0.52           & 5.12           & 0.04           & 0.15           & 0.02           & 0.05           & 0.27           \\ \hline
Size                  & $1024\times1024$ & $352\times352$ & $384\times384$ & $224\times224$ & $256\times256$ & $400\times300$ & $320\times320$ & $256\times256$ & $256\times256$ \\ \hline \hline
\multicolumn{1}{l|}{} & HRNet            & MINet          & CSF            & Amulet         & CPD            & F3N            & LDF            & ITSD           & GateNet        \\ \hline
Model Size(MB)        & 129.6            & 181.4          & 139.3          & 132.6          & 111.5          & 97.4           & 95.9           & 63.7           & -              \\ \hline
Time(s)               & 0.39             & 0.01           & 0.01           & 0.05           & 0.02           & 0.03           & 0.02           & 0.02           & 0.03           \\ \hline
Size                  & $1024\times1024$ & $320\times320$ & $224\times224$ & $256\times256$ & $352\times352$ & $352\times352$ & $352\times352$ & $288\times288$ & $384\times384$ \\ \hline
\end{tabular}}
\label{modelsize}
\end{table*}

\begin{table*}[!hbt]
\centering
\caption{Quantitative comparison with SOTA methods where the inputs are resized to high-resolution.}
\scalebox{0.68}{
\begin{tabular}{c|cccccc|cccccc}
\hline
                         & \multicolumn{6}{c}{HRSOD-TE}                                                                                                                                                             & \multicolumn{6}{c}{DAVIS-S}                                                                                                                                                             \\ \cline{2-13} 
\multirow{-2}{*}{Models} & $F_\beta^{max}$              & $F_\beta$                    & $S_m$                        & MAE                          & BDE                           & $B_\mu$                      & $F_\beta^{max}$              & $F_\beta$                    & $S_m$                        & MAE                          & BDE                           & $B_\mu$                      \\ \hline
CPD(High-Resolution)     & 0.868                        & 0.735                        & 0.809                        & 0.073                        & 181.770                       & 0.819                        & 0.720                        & 0.679                        & 0.799                        & 0.062                        & 126.281                       & 0.748                        \\
CPD(Low-Resolution)      & 0.876                        & 0.829                        & 0.887                        & 0.039                        & 72.686                        & 0.824                        & 0.878                        & 0.822                        & 0.903                        & 0.025                        & 36.649                        & 0.703                        \\ \hline\hline
EGNet(High-Resolution)   & 0.745                        & 0.693                        & 0.791                        & 0.082                        & 213.333                       & 0.867                        & 0.692                        & 0.644                        & 0.801                        & 0.069                        & 149.537                       & 0.821                        \\
EGNet(Low-Resolution)    & 0.883                        & 0.814                        & 0.888                        & 0.044                        & 73.500                        & 0.896                        & 0.886                        & 0.794                        & 0.897                        & 0.030                        & 37.369                        & 0.799                        \\ \hline\hline
F3N(High-Resolution)     & 0.834                        & 0.757                        & 0.825                        & 0.066                        & 187.942                       & 0.798                        & 0.698                        & 0.712                        & 0.826                        & 0.054                        & 130.603                       & 0.716                        \\
F3N(Low-Resolution)      & 0.900                        & 0.853                        & 0.897                        & 0.035                        & 65.901                        & 0.817                        & 0.915                        & 0.845                        & 0.913                        & 0.020                        & 45.106                        & 0.719                        \\ \hline\hline
GCPA(High-Resolution)    & 0.810                        & 0.771                        & 0.830                        & 0.066                        & 164.142                       & 0.793                        & 0.750                        & 0.714                        & 0.829                        & 0.057                        & 122.068                       & 0.708                        \\
GCPA(Low-Resolution)     & 0.889                        & 0.827                        & 0.894                        & 0.039                        & 70.320                        & 0.873                        & 0.912                        & 0.833                        & 0.924                        & 0.021                        & 24.132                        & 0.759                        \\ \hline\hline
MINet(High-Resolution)   & 0.687                        & 0.629                        & 0.742                        & 0.111                        & 250.149                       & 0.913                        & 0.580                        & 0.508                        & 0.681                        & 0.129                        & 176.671                       & 0.888                        \\
MINet(Low-Resolution)    & 0.902                        & 0.851                        & 0.903                        & 0.032                        & 76.291                        & 0.849                        & 0.915                        & 0.864                        & 0.926                        & 0.019                        & 32.304                        & 0.742                        \\ \hline\hline
LDF(High-Resolution)     & 0.650                        & 0.586                        & 0.673                        & 0.133                        & 208.545                       & 0.898                        & 0.590                        & 0.553                        & 0.696                        & 0.101                        & 150.540                       & 0.844                        \\
LDF(Low-Resolution)      & 0.905                        & 0.866                        & 0.905                        & 0.032                        & 58.655                        & 0.812                        & 0.911                        & 0.864                        & 0.922                        & 0.019                        & 35.496                        & 0.713                        \\ \hline\hline
CSF(High-Resolution)     & 0.802                        & 0.756                        & 0.843                        & 0.063                        & 181.705                       & 0.873                        & 0.700                        & 0.685                        & 0.824                        & 0.058                        & 137.592                       & 0.816                        \\
CSF(Low-Resolution)      & 0.894                        & 0.832                        & 0.900                        & 0.038                        & 71.293                        & 0.922                        & 0.899                        & 0.822                        & 0.912                        & 0.025                        & 30.488                        & 0.848                        \\ \hline\hline
Ours                     & {\color[HTML]{FE0000} 0.918} & {\color[HTML]{FE0000} 0.902} & {\color[HTML]{FE0000} 0.912} & {\color[HTML]{FE0000} 0.027} & {\color[HTML]{FE0000} 48.468} & {\color[HTML]{FE0000} 0.711} & {\color[HTML]{FE0000} 0.933} & {\color[HTML]{FE0000} 0.919} & {\color[HTML]{FE0000} 0.933} & {\color[HTML]{FE0000} 0.015} & {\color[HTML]{FE0000} 15.676} & {\color[HTML]{FE0000} 0.536} \\ \hline
\end{tabular}}
\label{size1024}
\end{table*}

\begin{table*}[!hbt]
\centering
\caption{Quantitative comparison with SOTA methods which are finetuned on HRSOD-Training dataset.}
\scalebox{0.73}{
\begin{tabular}{c|llllll|llllll}
\hline
                         & \multicolumn{6}{c|}{HRSOD-TE}                                                                                                                                                            & \multicolumn{6}{c}{DAVIS-S}                                                                                                                                                              \\ \cline{2-13} 
\multirow{-2}{*}{Models} & $F_\beta^{max}$              & $F_\beta$                           & $S_m$                           & MAE                          & BDE                           & $B_\mu$                         & $F_\beta^{max}$              & $F_\beta$                           & $S_m$                           & MAE                          & BDE                           & $B_\mu$                         \\ \hline
BASNet(finetune)         & 0.885                        & 0.836                        & 0.904                        & 0.035                        & 64.475                        & 0.813                        & 0.866                        & 0.838                        & 0.911                        & 0.023                        & 25.924                        & {\color[HTML]{000000} 0.659} \\ 
BASNet(original)         & 0.878                        & 0.831                        & 0.890                        & 0.038                        & 67.643                        & 0.823                        & 0.857                        & 0.806                        & 0.881                        & 0.039                        & 46.283                        & 0.705                        \\ \hline\hline
CPD(finetune)            & 0.890                        & 0.846                        & 0.899                        & 0.035                        & 80.857                        & {\color[HTML]{3166FF} 0.783} & 0.890                        & 0.871                        & 0.925                        & 0.020                        & 29.376                        & {\color[HTML]{3166FF} 0.671} \\ 
CPD(original)            & 0.876                        & 0.829                        & 0.887                        & 0.039                        & 72.686                        & 0.824                        & 0.878                        & 0.822                        & 0.903                        & 0.025                        & 36.649                        & 0.703                        \\ \hline\hline
EGNet(finetune)          & 0.890                        & 0.857                        & {\color[HTML]{32CB00} 0.911} & {\color[HTML]{3166FF} 0.031} & 69.084                        & 0.797                        & 0.899                        & {\color[HTML]{32CB00} 0.881} & 0.926                        & 0.021                        & 30.674                        & 0.686                        \\ 
EGNet(original)          & 0.883                        & 0.814                        & 0.888                        & 0.044                        & 73.500                        & 0.896                        & 0.886                        & 0.794                        & 0.897                        & 0.030                        & 37.369                        & 0.799                        \\ \hline\hline
GCPA(finetune)           & 0.895                        & 0.837                        & {\color[HTML]{FE0000} 0.912} & 0.032                        & 64.656                        & 0.846                        & 0.918                        & 0.857                        & 0.927                        & 0.019                        & {\color[HTML]{3166FF} 22.312} & 0.746                        \\ 
GCPA(original)           & 0.889                        & 0.827                        & 0.894                        & 0.039                        & 70.320                        & 0.873                        & 0.912                        & 0.833                        & 0.924                        & 0.021                        & 24.132                        & 0.759                        \\ \hline\hline
F3N(finetune)            & 0.905                        & {\color[HTML]{3166FF} 0.865} & 0.909                        & 0.033                        & 60.803                        & 0.787                        & {\color[HTML]{3166FF} 0.920} & 0.860                        & 0.921                        & 0.019                        & 29.106                        & {\color[HTML]{32CB00} 0.661} \\ 
F3N(original)            & 0.900                        & 0.853                        & 0.897                        & 0.035                        & 65.901                        & 0.817                        & 0.915                        & 0.845                        & 0.913                        & 0.020                        & 45.106                        & 0.719                        \\ \hline\hline
PFPN(finetune)           & 0.896                        & 0.840                        & 0.904                        & 0.038                        & {\color[HTML]{32CB00} 55.027} & 0.786                        & 0.901                        & 0.845                        & 0.920                        & 0.022                        & {\color[HTML]{32CB00} 21.388} & 0.728                        \\ 
PFPN(original)           & 0.889                        & 0.825                        & 0.897                        & 0.042                        & 65.048                        & 0.897                        & 0.886                        & 0.822                        & 0.912                        & 0.025                        & 30.488                        & 0.848                        \\ \hline\hline
ITSD(finetune)           & 0.834                        & 0.774                        & 0.863                        & 0.052                        & 117.554                       & 0.906                        & 0.820                        & 0.754                        & 0.873                        & 0.041                        & 75.461                        & 0.830                        \\ 
ITSD(original)           & 0.824                        & 0.715                        & 0.834                        & 0.071                        & 139.943                       & 0.924                        & 0.806                        & 0.687                        & 0.843                        & 0.055                        & 92.864                        & 0.861                        \\ \hline\hline
MINet(finetune)          & {\color[HTML]{3166FF} 0.908} & {\color[HTML]{32CB00} 0.871} & 0.908                        & 0.029                        & 66.089                        & {\color[HTML]{32CB00} 0.749} & {\color[HTML]{32CB00} 0.923} & {\color[HTML]{3166FF} 0.879} & 0.928                        & {\color[HTML]{32CB00} 0.017} & 25.408                        & 0.692                        \\ 
MINet(original)          & 0.902                        & 0.851                        & 0.903                        & 0.032                        & 76.291                        & 0.849                        & 0.915                        & 0.864                        & 0.926                        & 0.019                        & 32.304                        & 0.742                        \\ \hline\hline
LDF(finetune)            & {\color[HTML]{32CB00} 0.910} & 0.862                        & {\color[HTML]{3166FF} 0.910} & {\color[HTML]{3166FF} 0.031} & 77.098                        & 0.812                        & {\color[HTML]{3166FF} 0.920} & 0.867                        & 0.922                        & {\color[HTML]{3166FF} 0.018} & 42.226                        & 0.727                        \\ 
LDF(original)            & 0.905                        & 0.866                        & 0.905                        & 0.032                        & 58.655                        & 0.812                        & 0.911                        & 0.864                        & 0.922                        & 0.019                        & 35.496                        & 0.713                        \\ \hline\hline
GateNet(finetune)        & {\color[HTML]{32CB00} 0.910} & 0.856                        & 0.909                        & {\color[HTML]{32CB00} 0.029} & 76.434                        & 0.821                        & {\color[HTML]{32CB00} 0.923} & 0.872                        & {\color[HTML]{3166FF} 0.930} & 0.019                        & 36.984                        & 0.706                        \\ 
GateNet(original)        & 0.905                        & 0.825                        & 0.906                        & 0.035                        & 79.468                        & 0.886                        & 0.914                        & 0.825                        & 0.923                        & 0.023                        & 44.827                        & 0.778                        \\ \hline\hline
CSF(finetune)            & 0.902                        & 0.859                        & 0.909                        & {\color[HTML]{32CB00} 0.029} & {\color[HTML]{3166FF} 56.425} & 0.884                        & 0.910                        & 0.870                        & {\color[HTML]{32CB00} 0.931} & {\color[HTML]{32CB00} 0.017} & 24.669                        & 0.791                        \\ 
CSF(original)            & 0.894                        & 0.832                        & 0.900                        & 0.038                        & 71.293                        & 0.922                        & 0.899                        & 0.822                        & 0.912                        & 0.025                        & 30.488                        & 0.848                        \\ \hline\hline
Ours                     & {\color[HTML]{FE0000} 0.918} & {\color[HTML]{FE0000} 0.902} & {\color[HTML]{FE0000} 0.912} & {\color[HTML]{FE0000} 0.027} & {\color[HTML]{FE0000} 48.468} & {\color[HTML]{FE0000} 0.711} & {\color[HTML]{FE0000} 0.933} & {\color[HTML]{FE0000} 0.919} & {\color[HTML]{FE0000} 0.933} & {\color[HTML]{FE0000} 0.015} & {\color[HTML]{FE0000} 15.676} & {\color[HTML]{FE0000} 0.536} \\ \hline
\end{tabular}}
\label{refined}
\end{table*}

\subsection{Quantitative Comparison with different settings}
Although the effectiveness of our method has been confirmed by existing quantitative comparison experiments, to further illustrate the superiority of our method in handling high-resolution SOD task, we modify the setting of existing methods to allow for a more comprehensive comparison.

First, we change the input for the current SOTA methods from low-resolution (e.g., typical size $320\times320$, $352\times352$) to high-resolution ($1024\times1024$). The results are reported in Table.\ref{size1024}. It can be found that all the compared SOTA methods perform better at low-resolution on most evaluation metrics. Therefore, we only compare our methods to these SOTA methods' low-resolution results in our main paper. In particular, it is worth pointing out that due to GPU memory limitations, we cannot run BASNet, PFPN and ITSD at high-resolution. So we don't report their results in Table.\ref{size1024}.

Then, we fine-tune 11 SOTA methods on high-resolution datasets (HRSOD-Training) which have high quality annotations, the results are reported on Table.\ref{refined}. As can be seen, high annotation quality can improve their original performance. However, even fine-tuned on HRSOD-Training datasets, our method (only trained on DUTS) still outperforms all of them by a large margin. 

\begin{figure*}[!t]
\centering
\includegraphics[scale=0.53]{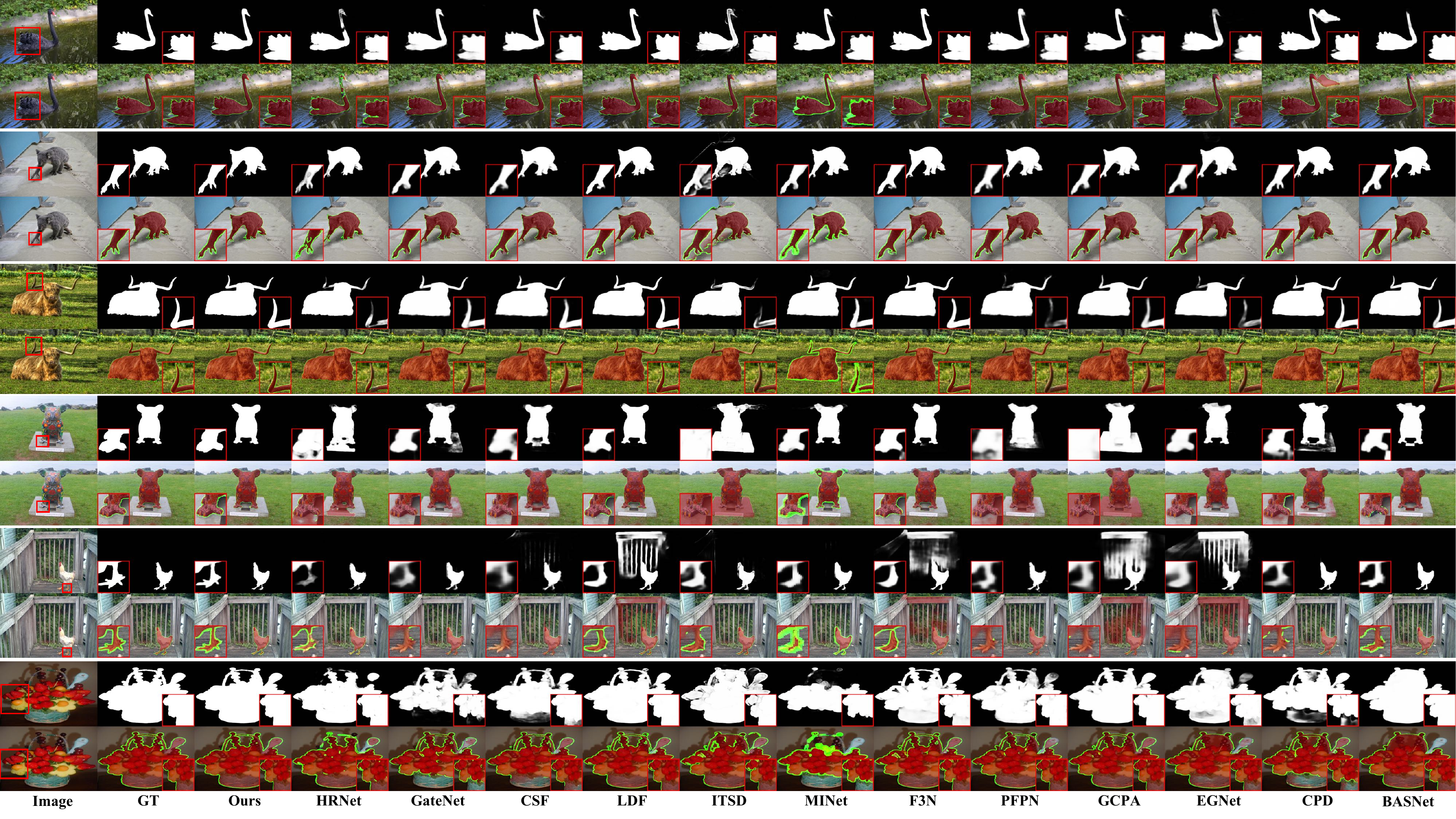}
\caption{Visual comparison between our method and other SOTA methods. Each sample occupies two rows. Best viewed by zooming in. It can be clearly observed that our method achieves impressive performance in all these cases.}
\label{Fig1}
\end{figure*}

\subsection{Qualitative Comparison}
As shown in Fig.\ref{Fig1}, we provide a comprehensive qualitative
comparison of our method with other 12 methods on challenging cases.
These visual examples can further demonstrate that our method is able
to restore accurate and complete boundaries of salient objects.

\begin{figure*}[!t]
\centering
\includegraphics[scale=0.65]{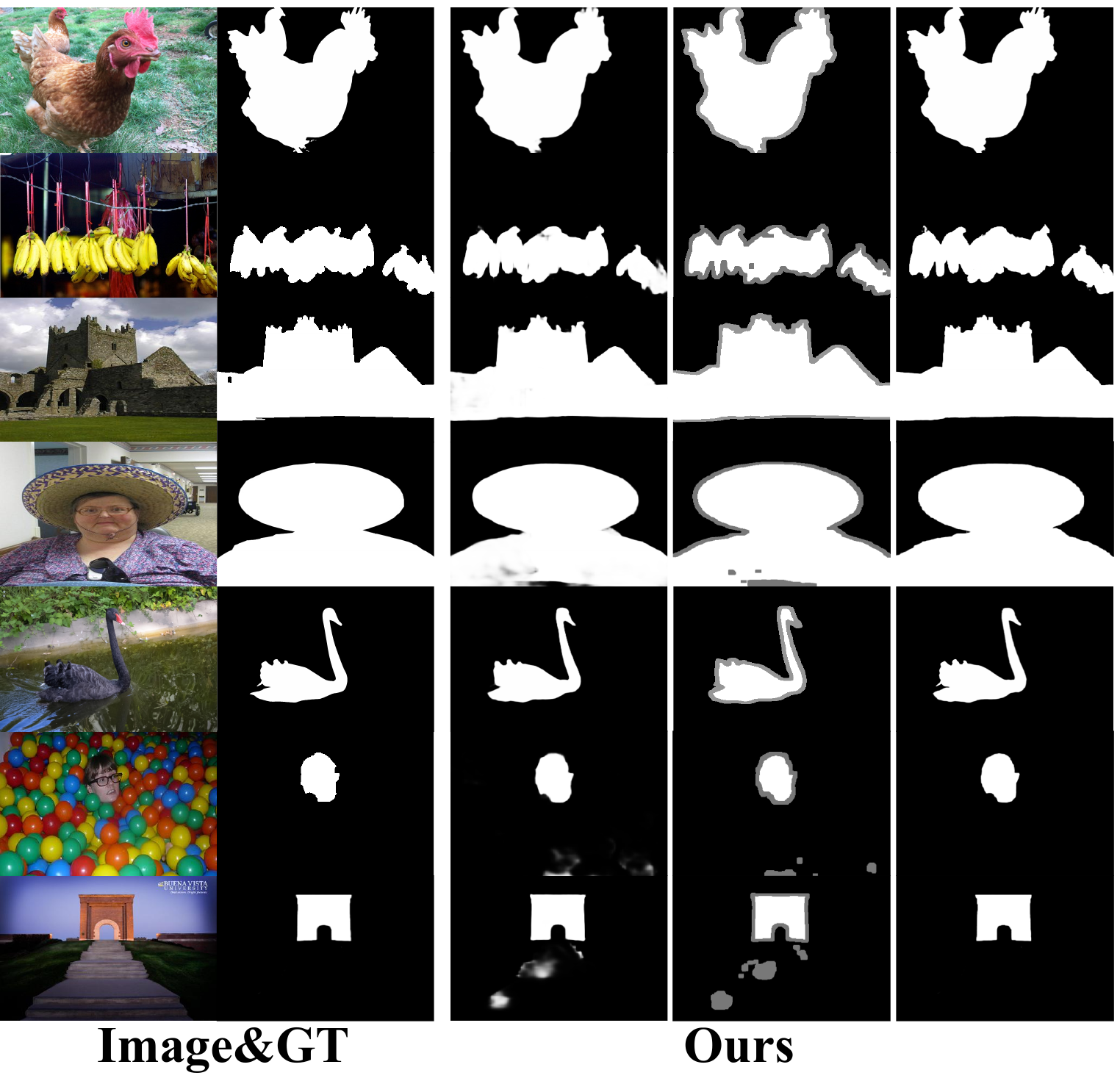}
\caption{Examples of coarse saliency maps, trimaps and refined saliency map.}
\label{trimaps}
\end{figure*}

\section{More analyses of the proposed disentangled framework}
\begin{table*}
\centering
\caption{Ablation Studies of disentangled framework.}
\scalebox{0.6}{
\begin{tabular}{c|cccccc|cccccc}
\hline
                                &                              & \multicolumn{3}{c}{HRSOD-TE}                                                               &                               &                              &                              & \multicolumn{3}{c}{DAIVS-S}                                                                &                               &                              \\ \cline{2-13} 
\multirow{-2}{*}{Conﬁgurations} & $F_\beta^{max}$              & $F_\beta$                    & $S_m$                        & MAE                          & BDE                           & $B_\mu$                      & $F_\beta^{max}$              & $F_\beta$                    & $S_m$                        & MAE                          & BDE                           & $B_\mu$                      \\ \hline
Regression-Regression                      & 0.912                        & 0.894                        & 0.899                        & 0.031                        & 56.251                        & 0.814                        & 0.923                        & 0.909                        & 0.918                        & 0.019                        & 22.737                        & 0.649                        \\
Classification-Classification                         & 0.913                        & 0.895                        & 0.898                        & 0.030                        & 54.143                        & 0.809                        & 0.921                        & 0.907                        & 0.921                        & 0.020                        & 23.892                        & 0.662                        \\
Ours                            & {\color[HTML]{FE0000} 0.918} & {\color[HTML]{FE0000} 0.902} & {\color[HTML]{FE0000} 0.912} & {\color[HTML]{FE0000} 0.027} & {\color[HTML]{FE0000} 48.468} & {\color[HTML]{FE0000} 0.711} & {\color[HTML]{FE0000} 0.933} & {\color[HTML]{FE0000} 0.919} & {\color[HTML]{FE0000} 0.933} & {\color[HTML]{FE0000} 0.015} & {\color[HTML]{FE0000} 15.676} & {\color[HTML]{FE0000} 0.536} \\ \hline
\end{tabular}}
\label{disentangled}
\end{table*}

As described, high-resolution salient object detection task should be disentangled into two tasks. One can be viewed as a classic classification task, while the other one is a typical regression task. To further illustrate the validity of our theory, we conduct additional experiments. Specifically, we consider these two tasks as regression or classification tasks simultaneously. The results are reported in Table.\ref{disentangled}. Compared with our proposed method, if we take the disentangled framework as the combination of the two regression or classification tasks, the performance will be degraded. Because the purpose of the proposed disentangled framework is to capture sufficient semantics at low-resolution (LRSCN Stage) and refine accurate boundary at high-resolution (HRRN Stage), which should be viewed as a classic classification task and a typical regression task. Fig.\ref{trimaps} shows some examples that our proposed HRRN can further refine accurate boundary, guided by trimaps. Specifically, column.3 and column.4 show the saliency maps and trimaps generated by LRSCN, and column.5 shows the results refined by HRRN. From Fig.\ref{trimaps}, guided by trimaps, our proposed HRRN can further refine the pixels value in uncertain regions to get more clear saliency results. 

Aforementioned work LDF~\cite{DBLP:conf/cvpr/WeiWWSH020}  has also introduced concepts related to decoupling. However, they still try to address the SOD task under a single regression framework.  Their approach is essentially an expansion of additional boundary supervision, which
barely touches the very nature of the SOD. As illustrated in our experiments, it is more natural to disentangle the SOD into two different tasks.

\section{Annotation Problems}
\begin{figure*}[!t]
\centering
\includegraphics[scale=0.65]{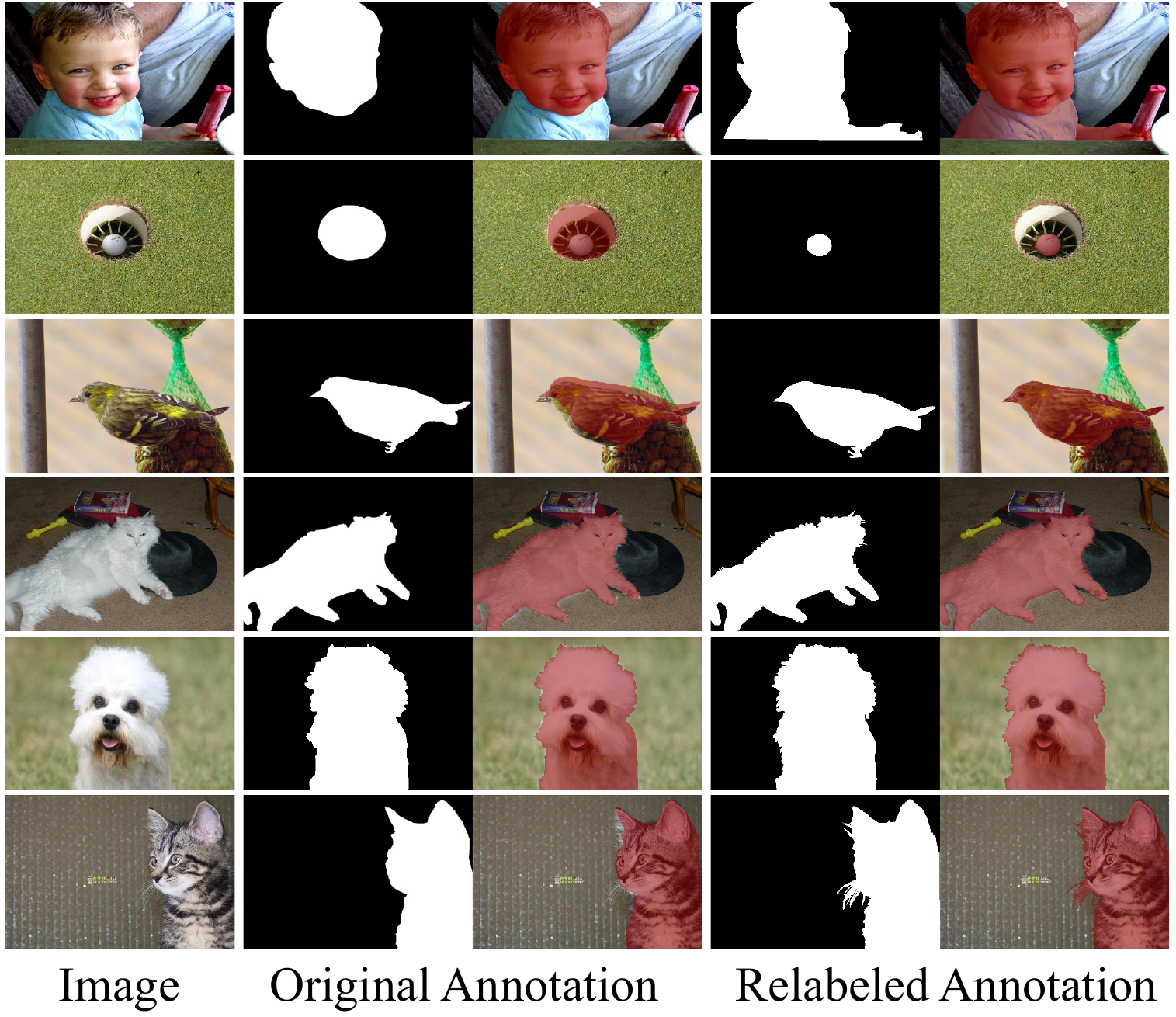}
\caption{Examples that have annotation quality problem. Best viewed by zooming in.}
\label{annoprob}
\end{figure*}

\begin{table*}[!hbt]
\centering
\caption{Ablation Studies of $L_{saliency}$.}
\scalebox{0.6}{
\begin{tabular}{l|cccccc|cccccc}
\hline
\multicolumn{1}{c|}{}                                &                              & \multicolumn{3}{c}{HRSOD-TE}                                                               &                               &                              &                              & \multicolumn{3}{c}{DAIVS-S}                                                                &                               &                              \\ \cline{2-13} 
\multicolumn{1}{c|}{\multirow{-2}{*}{Conﬁgurations}} & $F_\beta^{max}$              & $F_\beta$                    & $S_m$                        & MAE                          & BDE                           & $B_\mu$                      & $F_\beta^{max}$              & $F_\beta$                    & $S_m$                        & MAE                          & BDE                           & $B_\mu$                      \\ \hline
LRSCN($L_{trimap}$)+HRRN                               & 0.895                        & 0.870                        & 0.883                        & 0.035                        & 75.732                        & 0.879                        & 0.900                        & 0.880                        & 0.890                        & 0.026                        & 41.221                        & 0.733                        \\
LRSCN($L_P+L_{trimap}$)+HRRN                           & 0.912                        & 0.898                       & 0.908                       & 0.029                        & 53.040                        & 0.764                        & 0.925                        & 0.910                        & 0.926                        & 0.018                        & 19.022                        & 0.569                        \\
LRSCN($L_P+L_R+L_{trimap}$)+HRRN                       & 0.917                        & 0.900                        & 0.910                        & 0.029                        & 52.048                        & 0.743                        & 0.932                        & 0.914                        & 0.930                        & 0.017                        & 17.688                        & 0.552                        \\
LRSCN($L_P+L_R+L_O+L_{trimap}$)+HRRN                   & {\color[HTML]{FE0000} 0.918} & {\color[HTML]{FE0000} 0.902} & {\color[HTML]{FE0000} 0.912} & {\color[HTML]{FE0000} 0.027} & {\color[HTML]{FE0000} 48.468} & {\color[HTML]{FE0000} 0.711} & {\color[HTML]{FE0000} 0.933} & {\color[HTML]{FE0000} 0.919} & {\color[HTML]{FE0000} 0.933} & {\color[HTML]{FE0000} 0.015} & {\color[HTML]{FE0000} 15.676} & {\color[HTML]{FE0000} 0.536} \\ \hline
\end{tabular}}
\label{table1}
\end{table*}

As described in~\cite{DBLP:conf/iccv/ZengZLZL19}, widely used saliency datasets have some problems in annotation quality. So, to quantify the annotation quality problem, we randomly select 100 images from DUT-TR, and 10 of them have easily spotted annotation errors. We manually relabel the 10 images. The $B_\mu$ between the two different annotations is 0.49 and 42\% of the boundary pixel annotations are inaccurate. Fig.\ref{annoprob} shows some examples which have annotation problems, including wrong semantic annotation (row 1 and row 2), boundary annotation shifting (row 3) and low contour accuracy (row4, row5 and row 6). In conclusion, the DUTS-TR training dataset does have annotation problems~\cite{DBLP:conf/iccv/ZengZLZL19}, and we relabeled some examples to demonstrate these problems in the supplemental material. Since correcting annotations for the whole DUT-TR is a time-consuming task, we will provide an accurate GT of DUT-TR in the future for statistical analysis

\section{Details of $L_{saliency}$}
As described, to guarantee the arruracy of trimap, we add extra saliency supervision $L_{saliency}$ as the supplement of trimap supervision. Here we give more details about $L_{saliency}$.

After LRSCN, the prediction saliency map is $S$, and the binary groundtruth is $G$. In SOD, binary cross entropy (BCE) is the most widely used loss function, and it is a pixel-wise loss which is defined as:
\begin{equation}
\begin{aligned}
L_{Pixel} = -( Glog(S) + (1-G)log(1-S)   ).
\end{aligned}
\end{equation}

To learn the structural information of the salient objects, following the setting of~\cite{DBLP:journals/tip/WangBSS04,DBLP:conf/iccv/FanCLLB17},
we use the sliding window fashion to model region-level similarity between groundtruth and saliency map. The corresponding regions are denoted as $S_i = \{S_i:i=1,...M\}$ and $G_i = \{G_i:i=1,...M\}$, where $M$ is the total number of region. Then we use SSIM to evaluate the similarity between $S_i$ and $G_i$, which is defined as:
\begin{equation}
SSD_i = \frac{ (2\mu_s\mu_g+C_1)(2\sigma_{sg}+C_2)  }{ (\mu_s^2+\mu_g^2+C_1)(\sigma_s^2+\sigma_g^2+C_2)}
\end{equation}
where local statistics $\mu_s$, $\sigma_s$ is mean and std vector of $S_i$, $\mu_g$, $\sigma_g$ is mean and std vector of $G_i$. The overall loss function is defined as:
\begin{equation}
L_{Region} = 1 - \frac{1}{M}\sum_{i=1}^{M}SSD_i.
\end{equation}

Finally, inspired by~\cite{DBLP:conf/iccv/ZhaoGWC19}, we directly optimize the F-measure to learn the global information from groundtruth. For easy remembering, we denote F-measure as $F_\beta$ in the following. $F_\beta$ is defined as:
\begin{equation}
precision = \frac{ \sum{ S \cdot G }}{ \sum{S} + \epsilon}, \ \ recall = \frac{ \sum{ S \cdot G }}{ \sum{G} + \epsilon},
\end{equation}
\begin{equation}
F_\beta = \frac{ (1+\beta^2) \cdot precision \cdot recall }{ \beta^2 \cdot precision + recall },
\end{equation}
where $\cdot$ means pixel-wise multiplication, $\epsilon=1e^{-7}$ is a regularization constant to avoid division of zero. $L_{Object}$ loss function is defined as:
\begin{equation}
L_{Object} = 1 -  F_\beta.
\end{equation}
The whole loss is defined as:
\begin{equation}
L = L_{Object} + L_{Region} + L_{Pixel}.
\end{equation}
Besides, following~\cite{DBLP:conf/cvpr/QinZHGDJ19,DBLP:journals/corr/abs-1911-11445}, we used multi-levels saliency supervision to facilitate sufficient training, so the whole saliency loss is defined as:
\begin{equation}
L_{saliency} = \sum_{i=1}^{4}\frac{1}{2^{i-1}}L_{i},
\end{equation}
where $i$ means the i-th level.

To further validate the role of $L_{saliency}$, we train the LRSCN with different loss functions and the results are reported on Table.\ref{table1}. As can be can, without $L_{saliency}$, the performance is dropped lot. Because the trimap groundtruth is randomly generated from binary groundtruth, so only using $L_{trimap}$ cannot maintain consistency between trimap and saliency map. When we only add $L_P$ on multi-levels, the model can already achieve the largest performance boost. A better performance has been achieved through the combination of $L_P$, $L_R$ and $L_O$.

\begin{figure}[!t]
\centering
\includegraphics[scale=0.7,width=8.5cm]{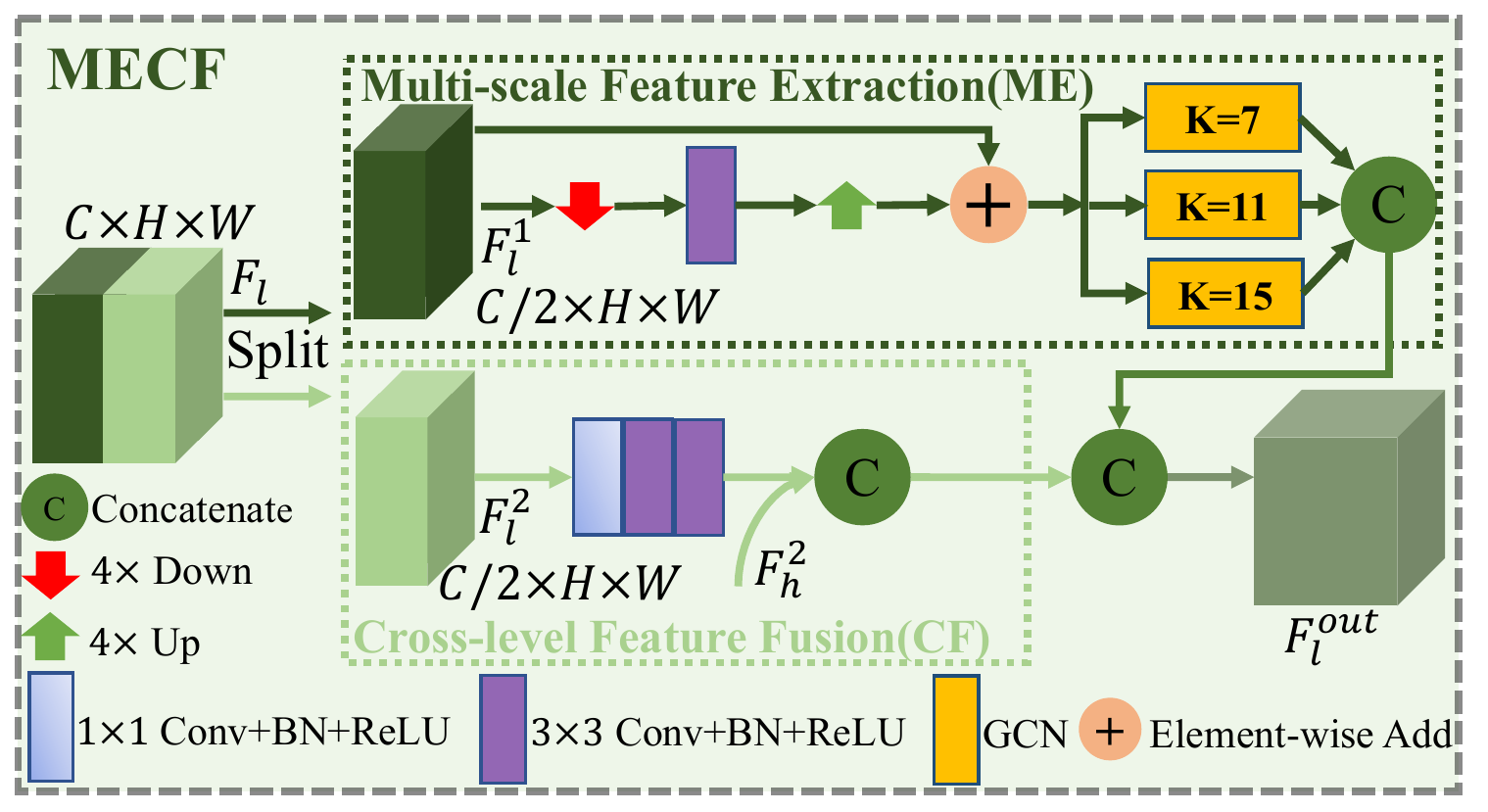}
\caption{Architecture of MECF Module.}
\label{supp-MECF}
\end{figure}

\section{Details of MECF Module}
As described, we develop a multi-scale feature extraction module (ME) and cross-level feature fusion module (CF) to help LRSCN capture sufficient semantics at low-resolution. Here we give more details about MECF module. The architecture of MECF Module is shown in Fig.\ref{supp-MECF}.

Multi-scale feature extraction module can allow each spatial location to view the local context at small scale spaces and capture multi-scale contextual information, which can enlarge the feature $F^1_l$ receptive field. Specifically, we first use an average pooling and a $3 \times 3$ convolutional layer to downsample $F^1_l$. Then upsampled feature from small scale is added with $F^1_l$. Finally, Global Convolutional Network (GCN)~\cite{DBLP:conf/cvpr/PengZYLS17} is used to further enlarge the feature receptive field. Because $F_3^1$ and $F_4^1$ are close to the input and receptive field is relatively small, we use GCNs with $k = 7,11,15$ to fully enlarge receptive field. Receptive fields of $F_5^1$ and $F_6^1$ are relatively bigger, we only use GCNs with $k=7,11$ and $k = 7$.

Low-level features have rich details but full of background noises, so we design cross-level feature fusion module, which can leverage the rich semantics of high-level feature $F^2_h$ and help restrain the non-salient regions in low-level features. Specifically, we first use a $1 \times 1$ convolutional layer to compress the channels of $F^2_l$, then use two $3 \times 3$ convolutional layer to transfer the feature for SOD task. Finally, the transferred feature is fused with high-level feature $F^2_h$ as the output of this module. Each of these convolution layers is followed by a batch normalization~\cite{DBLP:conf/icml/IoffeS15} and a ReLU activation~\cite{DBLP:conf/nips/HahnloserS00}. 

\section{Formulas of Evaluation Metrics}
Following ~\cite{DBLP:conf/iccv/ZengZLZL19} and ~\cite{DBLP:conf/cvpr/ZhangYLSLD20}, we use Boundary Displacement Error(BDE)~\cite{DBLP:conf/eccv/FreixenetMRMC02} and $B_\mu$ metrics to evaluate the boundary quality. 

BDE measures the average displacement error of boundary pixels between two predictions, which can be formulated as:
\begin{equation}
BDE(X,Y)=\frac{\sum_x inf_{y \in Y}d(x,y)}{2N_X} + \frac{\sum_y inf_{x \in X}d(x,y)}{2N_Y}, 
\end{equation}
where $X$ and $Y$ are two boundary pixel sets which represent saliency prediction and their corresponding groundtruth, and $x$, $y$ are pixels in them. $N_x$ and $N_y$ denote the number pixels in $X$ and $Y$. $inf$ represents for the infimum and $d(\cdot)$ denotes Euclidean distance.

$B_\mu$ evaluates the structure alignment between saliency map and their groundtruth, it can be expressed as: 
\begin{equation}
B_\mu = 1-\frac{2\sum(g_sg_y)}{\sum(g^2_s+g^2_y)}, 
\end{equation}
where $g_s$ and $g_y$ represent the binarized edge maps of predicted saliency map and groundtruth. Following~\cite{DBLP:conf/cvpr/ZhangYLSLD20}, we use Canny edge detector to compute edge maps. $B_\mu$ reﬂect the sharpness of predictions which is consistent with human perception.

\end{appendices}

\clearpage

{\small
\bibliographystyle{ieee_fullname}
\bibliography{egbib}
}

\end{document}